\pdfoutput=1

\documentclass[11pt]{article}

\usepackage[final]{acl}

\usepackage{times}
\usepackage{latexsym}
\usepackage[T1]{fontenc}

\usepackage[utf8]{inputenc}

\usepackage{microtype}

\usepackage{inconsolata}

\usepackage{graphicx}
\usepackage{booktabs}
\usepackage{multirow}
\usepackage{makecell}
\usepackage{amssymb}
\usepackage{amsmath}
\usepackage{amsfonts}
\usepackage{algorithm}
\usepackage{algpseudocode}
\usepackage{etoolbox}
\usepackage{float}
\algrenewcommand\algorithmicrequire{\textbf{Input:}}  
\algrenewcommand\algorithmicensure{\textbf{Output:}}

\title{Optimal Transport Depth Up-Scaling}

\author{
Mingzi Cao \quad Xi Wang \quad Nikolaos Aletras \\
School of Computer Science, University of Sheffield, United Kingdom \\
\texttt{\{mcao20,xi.wang,n.aletras\}@sheffield.ac.uk}
}

\begin{document}
\maketitle
\begin{abstract}

Pre-training Large Language Models (LLMs) from scratch at larger scales yields remarkable performance but incurs substantially high training costs. Depth up-scaling provides an efficient alternative by inserting new layers into a pre-trained LLM, avoiding training from scratch. However, most existing methods copying or averaging base layers for new layer, which misalign functionally corresponding neurons, leading to neuron permutation mismatch that harms performance. To address this issue, we propose Optimal Transport Depth Up-Scaling (OT-DUS), which leverages Optimal Transport (OT) theory to align and fuse functionally corresponding neurons module by module in adjacent base layers for new layer construction. OT-DUS achieves better overall performance in both general and specialized domains than existing methods for continual pre-training and supervised fine-tuning across different model sizes and model families. Our analysis of insertion strategies further finds that inserting new layers at higher positions yields not only stronger performance but also improved training efficiency.\footnote{Our code is available at \url{https://github.com/voalmciaf/OpT-DeUS}.}

\end{abstract}

\section{Introduction}

\begin{figure}[!ht]
\begin{center}
\includegraphics[width=0.9\columnwidth]{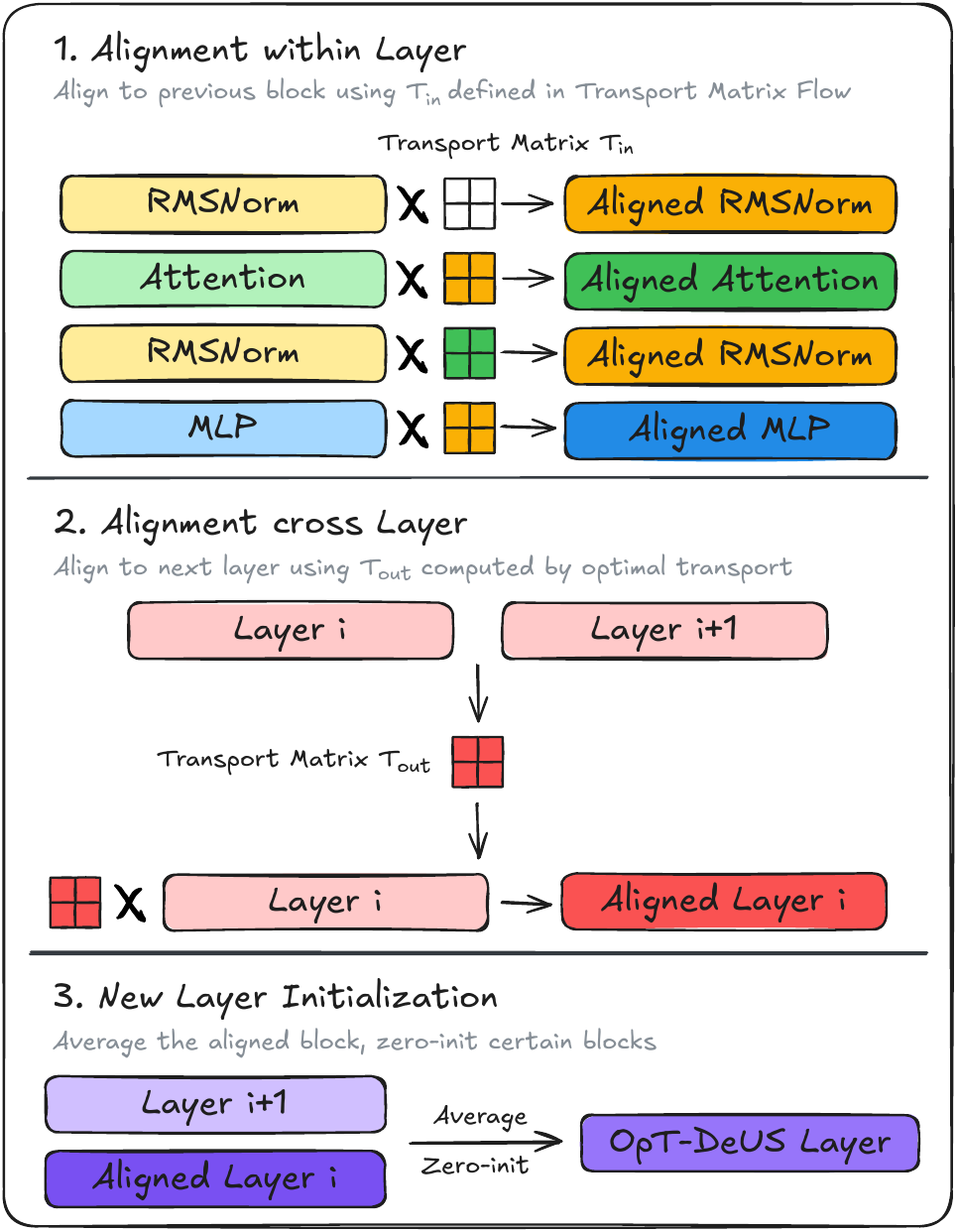}
\end{center}
  \caption{Illustration of our proposed OT-DUS. OT-DUS uses optimal transport to initialize new layers, each derived from two adjacent base layers $f_i$ and $f_{i+1}$. It first aligns each module $b$ to previous module $b-1$ in $f_i$ (i.e. Alignment within Layer), then aligns it to $b$ in $f_{i+1}$ (i.e. Alignment across Layer).}
  \label{fig:illu}
\end{figure}

Large Language Models (LLMs) performance is largely attributed to scaling laws, where capabilities often improve with increased model and data size \citep{NEURIPS2020_1457c0d6,kaplan2020scalinglawsneurallanguage,wei2022emergent,JMLR:v25:23-0870}. However, scaling poses sustainability challenges, stemming from increased computational \citep{thompson2022computationallimitsdeeplearning,MLSYS2022_462211f6,JMLR:v24:23-0069} and data \citep{10.1145/3442188.3445922,NEURIPS2023_42f22550, villalobos2024rundatalimitsllm} constraints. To address these challenges, model expansion approaches have been proposed, which increase the parameter size of a pre-trained model without altering its original architecture. This can be achieved by increasing the number of neurons per layer (i.e., width up-scaling) or the number of layers (i.e., depth up-scaling).

Depth up-scaling has shown better training efficiency and downstream performance than width up-scaling \citep{NEURIPS2024_143ea4a1}. Recent depth up-scaling approaches update only the new layers for training efficiency while mitigating catastrophic forgetting \citep{kim-etal-2024-solar,yang-etal-2025-lesa}. Depth up-scaling largely relies on copying \citep{wu-etal-2024-llama,kim-etal-2024-solar,NEURIPS2024_143ea4a1} or averaging \citep{yano-etal-2025-step} for new layer construction. While effective, directly copying or averaging from base layers ignores neuron permutation differences across layers, which can incorrectly align neurons, leading to neuron permutation mismatches that harm performance \citep{pmlr-v44-li15convergent,NEURIPS2019_ecb287ff,pmlr-v97-yurochkin19a}. An alternative approach \citep{yang-etal-2025-lesa} initializes new layers by training an auxiliary neural network, but it introduces additional computational overhead that increases rapidly with model scale. These challenges motivate our main research question: \textit{How can new layers in depth up-scaling be effectively initialized to mitigate neuron permutation mismatches without substantial additional computation?}

Inspired by applying Optimal Transport (OT) \citep{NEURIPS2020_fb269786, imfeld2024transformer}, we propose Optimal Transport Depth Up-Scaling (OT-DUS). As shown in Figure~\ref{fig:illu}, OT-DUS aligns and fuses adjacent layers module-wise to create neuron-aligned new layers. Newly added layers are initialized via OT and inserted into the top half of the base model. Certain module weights are set to zero for better neuron alignment and function preservation. Our contributions are as follows:

\begin{itemize}
\item We introduce OT-DUS, which creates intermediate layer from adjacent layers by neuron alignment via OT. Experiments show that OT-DUS outperforms existing baselines on both continual pre-training and supervised fine-tuning training stages across various model sizes, model families, and diverse tasks.
\item OT-DUS achieves the highest overall efficiency, requiring minimal additional computation. Our comprehensive study on layer interpolation position shows that inserting new layers at higher positions leads to higher training efficiency while obtaining better performance.
\item Extensive ablation studies validate the contributions of the OT-DUS components, confirming that our design effectively mitigates neuron permutation mismatches. Analysis of neuron functionality  reveals a strong correlation between depth up-scaling performance and high transport matrix entropy.
\end{itemize}

\section{Related Work}
\subsection{Model Expansion} 

Model expansion accelerates neural network training by expanding a base model to reduce training time and computational overhead \citep{Chen2016Net2Net,pmlr-v48-wei16,chang2018multilevel,rusu2022progressiveneuralnetworks}. Network architecture preservation has proven effective for iterative expansion in encoder-only LLMs \citep{pmlr-v97-gong19a,yang2020progressivelystacking20multistage,chen-etal-2022-bert2bert}. More recently, various model expansion approaches have been explored for decoder-only LLMs. \citet{NEURIPS2024_143ea4a1} showed depth up-scaling yields greater training efficiency and stronger downstream performance compared to width up-scaling. However, prior work primarily focuses on expansion during the pre-training stage with a relatively large pre-training corpus \citep{pmlr-v162-shen22f,wang2023learning,wang2024lemon,yao2024masked,yano-etal-2025-step}, resulting in high overall computational costs. Limited work focuses on post-training expansion  \citep{kim-etal-2024-solar,wu-etal-2024-llama,yang-etal-2025-lesa}, using a substantially smaller corpus compared to the original pre-training corpus for training efficiency.

\subsection{Depth Up-Scaling} 
\noindent \textbf{Stacking.} Stacking methods insert a series of new layers, typically on top by copying layers from the base model \citep{NEURIPS2024_143ea4a1,kim-etal-2024-solar}. \citet{NEURIPS2024_143ea4a1} proposed stacking entire base layers for stronger downstream performance during pre-training. \citet{kim-etal-2024-solar} introduced SOLAR, a partial stacking approach that omits bottom and top base layers for continual pre-training. However, stacking requires updating the entire model, incurring extra computational costs. 

\begin{figure*}[!ht]
\begin{center}
\includegraphics[width=0.6\textwidth]{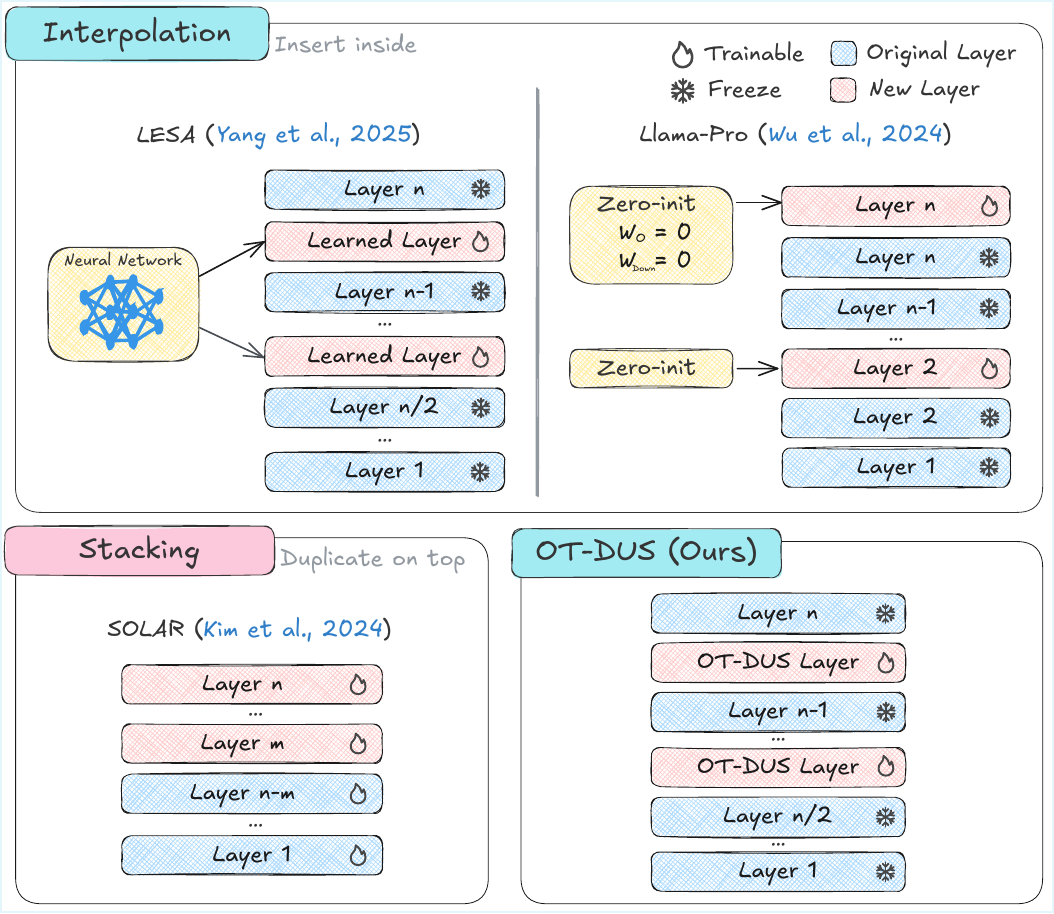}
\end{center}
  \caption{State-of-the-art depth up-scaling methods and our proposed OT-DUS.}
  \label{fig:main}
\end{figure*}

\noindent \textbf{Interpolation.} Interpolation methods insert new layers inside the base model. Previous work focuses on creating function preservation layers, where the expanded model performs identically to the base model before further training. Achieving function preservation leads to steadier learning processes and better performance. This is achieved by setting the LayerNorm weights to zero for new layer initialization \citep{pmlr-v162-shen22f}, initializing the entire new layer to zero \citep{wang2024lemon}, or employing dynamic masking mechanisms \citep{yao2024masked}. \citet{wu-etal-2024-llama} proposed LLaMA PRO, which initializes the inserted new layers by copying weights from the base model. For function preservation, the output matrices of attention and MLP in these new Transformer layers are set to zero, termed zero-init. \citet{yano-etal-2025-step} initialized new layers by averaging weights from adjacent base layers for pre-training. LESA \citep{yang-etal-2025-lesa} initializes new layers using an auxiliary network given adjacent layers at interpolation positions as input. However, existing methods largely rely on copying \citep{kim-etal-2024-solar, wu-etal-2024-llama} or averaging \citep{yano-etal-2025-step} to initialize new layers, neglecting neuron permutation differences.

\section{Preliminaries}

\subsection{Depth Up-scaling}

Let $\mathcal{M}$ be a \textit{base} LLM with $n$ Transformer layers $\{f_i\}_{i=1}^n$, parametrized by $\theta$.
The aim is to obtain an \textit{expanded} model $\mathcal{M}'$ with parameters $\theta'$ by introducing $k$ additional Transformer layers $\{f'_i\}_{i=1}^k$.
$\mathcal{M}'$ retains the same layer type and hidden size.

Each Transformer layer $f_i$ is composed of a sequence of modules.
We denote the parameters of layer $f_i$ by
$\{ \mathbf{W}_b^{(i)} \}_{b=1}^{B}$,
where $\mathbf{W}_b^{(i)}$ represents the weight matrix of module $b$ in layer $f_i$.
Accordingly, each new layer $f'_i$ is parameterized by
$\{ \mathbf{W'}_b^{(i)} \}_{b=1}^{B}$.

\noindent{\textbf{Stacking.}
$\mathcal{M}$ is expanded by adding a set of new layers on top of the base layers to obtain $\mathcal{M}'$.
$\circ$ denotes the connection between Transformer layers:
\begin{equation*}
\small
\mathcal{M}'(x; \theta') = f'_k \circ \cdots \circ f'_1 \circ f_n \circ \cdots \circ f_1(x).
\end{equation*}}

Each new layer $f'_i$ is initialized by duplicating a selected base layer $f_j$
\citep{NEURIPS2024_143ea4a1,kim-etal-2024-solar},
where $j$ is not necessarily equal to $i$:

\begin{equation*}
\small
\mathbf{W'}_b^{(i)} \leftarrow \mathbf{W}_b^{(j)}, \quad \forall b \in \{1,\dots,B\}.
\end{equation*}

\noindent{\textbf{Interpolation.}
Figure~\ref{fig:main} illustrates different interpolation strategies adopted by existing depth up-scaling methods.
$\mathcal{M}$ is expanded by inserting new layers between base layers as follows:
\begin{equation*}
\small
\mathcal{M}'(x; \theta') =
\begin{cases}
f'_i \circ f_i, & \text{if a new layer is inserted}, \\
f_i, & \text{otherwise}.
\end{cases}
\end{equation*}}

Interpolation methods initialize $f'_i$ from adjacent base layers
$f_i$ and $f_{i+1}$.
This corresponds to initializing each module weight matrix
$\mathbf{W'}_b^{(i)}$ using $\mathbf{W}_b^{(i)}$ and $\mathbf{W}_b^{(i+1)}$.
Prior work initialize $f'_i$ by copying
\citep{wu-etal-2024-llama,kim-etal-2024-solar,NEURIPS2024_143ea4a1},
averaging \citep{yano-etal-2025-step},
or prediction via an auxiliary network \citep{yang-etal-2025-lesa}:
\begin{equation*}
\small
\mathbf{W'}_b^{(i)} \leftarrow
\begin{cases}
\mathbf{W}_b^{(i)}, & \text{copying}, \\
\text{Avg}\!\left(\mathbf{W}_b^{(i)}, \mathbf{W}_b^{(i+1)}\right), & \text{averaging}, \\
\text{NN}\!\left(\mathbf{W}_b^{(i)}, \mathbf{W}_b^{(i+1)}\right), & \text{prediction}, \\
\text{OT-DUS}\!\left(\mathbf{W}_b^{(i)}, \mathbf{W}_b^{(i+1)}\right), & \text{our method},
\end{cases}
\end{equation*}

\subsection{Optimal Transport}

\label{sec:OT}
\begin{figure*}[!ht]
\begin{center}

\includegraphics[width=0.7\textwidth]{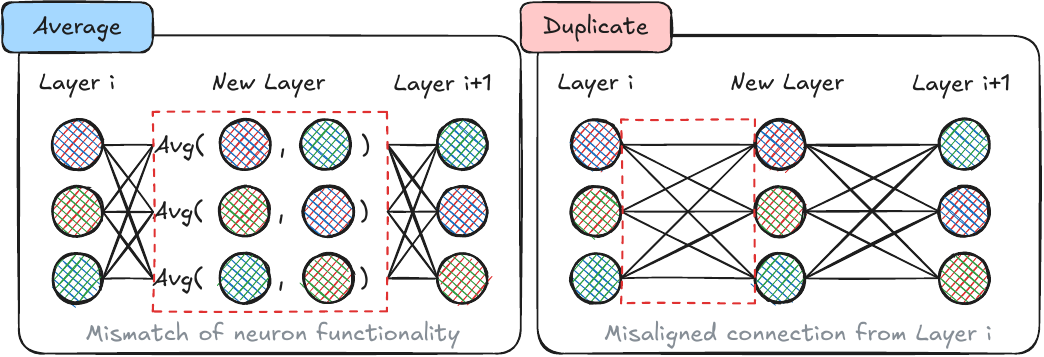}
\end{center}
  \caption{Neuron permutation mismatch caused by average and duplicate. Each column represents the order of neurons within layer. Multiple colours of each neuron represent multiple functions it contributes to. 
}
  \label{fig:Motivation}
\end{figure*}
Optimal Transport is a mathematical framework determining the most cost-effective way to transform one probability distribution into another, given a defined transport cost. Formally, let $\mu = \sum_{i=1}^n \alpha_i \delta(x^{(i)})$ and $\nu = \sum_{j=1}^m \beta_j \delta(y^{(j)})$ be two discrete probability measures supported on $\{x^{(i)}\}_{i=1}^n$ and $\{y^{(j)}\}_{j=1}^m$ with probability distribution $\boldsymbol{\alpha} = (\alpha_1, \dots, \alpha_n)$ and $\boldsymbol{\beta} = (\beta_1, \dots, \beta_m)$, respectively, where $\delta(s)$ denotes the unit mass at point $s$. Given a cost matrix $\mathbf{C} \in \mathbb{R}^{n \times m}$ where $\mathbf{C}_{ij}$ is the cost of transporting mass from $x^{(i)}$ to $y^{(j)}$, the Optimal Transport (OT) problem seeks an transport matrix that minimizes the total cost:

\begin{equation*}
\small
\begin{aligned}
\text{OT}(\mu,\nu, \mathbf{C})
&= \operatorname*{arg\,min}_{\mathbf{T} \in \mathbb{R}_+^{n \times m}} \sum_{i,j} \mathbf{T}_{ij} \mathbf{C}_{ij} \\
\text{s.t.} \: 
\mathbf{T} \mathbf{1}_m &= \boldsymbol{\alpha}, 
\mathbf{T}^\top \mathbf{1}_n = \boldsymbol{\beta}.
\end{aligned}
\end{equation*}

\section{Optimal Transport Depth Up-Scaling }
\label{sec:OT-DUS}

\subsection{Motivation}
\label{sec:motivation}

Copying and averaging from original layers are commonly used methods for creating new layers \citep{wu-etal-2024-llama, kim-etal-2024-solar,NEURIPS2024_143ea4a1, yano-etal-2025-step}. However, such approach neglects neuron permutation mismatch, which is widely present in deep neural networks and Transformers \citep{pmlr-v44-li15convergent,NEURIPS2019_ecb287ff,pmlr-v97-yurochkin19a}. During training, a single neuron may contribute to multiple functions \citep{nguyen2016multifacetedfeaturevisualizationuncovering}, and same-index neurons in different layers may not be functionally corresponding \citep{sajjad-etal-2022-neuron,10.1145/3728458}. As shown in Figure \ref{fig:Motivation}, averaging from base layers to initialize $f'_i$ can incorrectly merge neurons with different functionalities, while copying from $f_i$ to initialize $f'_i$ breaks the connectivity between $f_i$ and $f'_i$. Thus, directly copying or averaging  can cause misalignment between $f_i$ and $f_{i+1}$, potentially harming performance \citep{pmlr-v44-li15convergent,NEURIPS2019_ecb287ff,pmlr-v97-yurochkin19a}.

Neuron permutation mismatch can be mitigated by aligning neurons between $f_i$ and $f_{i+1}$ using OT, which models functional similarity per neuron. \citet{NEURIPS2020_fb269786} and \citet{imfeld2024transformer} show that aligning neurons layer-wise via OT leads to better-initialized new layers for model merging. Recent research further shows that using information from adjacent layers provides stronger initialization than random initialization in depth up-scaling \citep{NEURIPS2024_143ea4a1,yano-etal-2025-step,yang-etal-2025-lesa}. This inspires our Optimal Transport Depth Up-scaling (OT-DUS), illustrated in Figure~\ref{fig:illu}. OT-DUS is a progressive interpolation approach that updates only $f'$. It aligns and fuses layers $f_i$ and $f_{i+1}$ module by module to create $f'_i$ via OT. OT-DUS inserts new layers $f'_i$ in the top half of $\mathcal{M}$, between base layers $f_i$ and $f_{i+1}$. This layer interpolation strategy provides better performance (Section~\ref{sec:IP}) and training efficiency (Section~\ref{TE}).

\subsection{Transport Matrix Flow for OT-DUS}
\label{sec:TMF}

OT-DUS relies on two types of transport matrices: $\mathbf{T}_\mathbf{in}$ and $\mathbf{T}_\text{out}$. Each module weight matrix $\mathbf{W}_{b}^{(i)}$ in $f'_i$ is assigned a $\mathbf{T}_\text{in}$. $\mathbf{T}_\text{in}$ aligns $\mathbf{W}_{b}^{(i)}$ to $\mathbf{W}_{b-1}^{(i)}$ within the layer. $\mathbf{T}_\text{out}$ aligns $\mathbf{W}_{b}^{(i)}$ to $\mathbf{W}_{b}^{(i+1)}$ across layers. $\mathbf{T}_\text{in}$ for $\mathbf{W}_{b}^{(i)}$ is initialized by reusing the $\mathbf{T}_\text{out}$ from the previous module $\mathbf{W}_{b-1}^{(i)}$. $\mathbf{T}_\text{out}$ is computed by solving an OT problem (Section~\ref{sec:weight_init}).

\begin{table}[ht]
\centering

\setlength{\tabcolsep}{4pt}

\resizebox{\columnwidth}{!}{
\begin{tabular}{cccccccccc}
\toprule
& \multicolumn{2}{c}{\textbf{Normalization}}
& \multicolumn{4}{c}{\textbf{Attention}}
& \multicolumn{3}{c}{\textbf{MLP}} \\

\cmidrule(lr){2-3}
\cmidrule(lr){4-7}
\cmidrule(lr){8-10}

\textbf{Transport Matrix}
& \textbf{Pre-Attn}
& \textbf{Pre-MLP}
& \textbf{Query}
& \textbf{Key}
& \textbf{Value}
& \textbf{Output}
& \textbf{Gate}
& \textbf{Up}
& \textbf{Down} \\

\midrule
$\mathbf{T}_{\text{in}}$
& $\mathbf{I}$
& $\frac{1}{2}(\mathbf{T}_O + \mathbf{I})$
& $\mathbf{I}$
& $\mathbf{I}$
& $\mathbf{I}$
& $\underline{\mathbf{I}}$
& \underline{$\mathbf{T}_O$}
& \underline{$\mathbf{T}_O$}
& $\underline{\mathbf{I}}$ \\

$\mathbf{T}_{\text{out}}$
& $\mathbf{I}$
& $\frac{1}{2}(\mathbf{T}_O + \mathbf{I})$
& $\mathbf{T}_Q$
& $\mathbf{T}_K$
& $\mathbf{T}_V$
& $\mathbf{T}_O$
& $\mathbf{T}_{\text{Gate}}$
& $\mathbf{T}_{\text{Up}}$
& $\mathbf{T}_{\text{Down}}$ \\

\bottomrule
\end{tabular}
}

\caption{Transport Matrix Flow.  We manually set $\mathbf{T}_\text{in}$ to each module for alignment within layer. $\mathbf{T}_\text{out}$ is calculated through OT for alignment across layers (except Normalization component). Modules highlighted in \underline{underline} deviate from \citet{imfeld2024transformer}.}
\label{fig:T_Flow}
\end{table}

 We use Transport Matrix Flow ($\text{TMF}$) to define the assignment of $\mathbf{T}_\text{in}$ for each module (Table~\ref{fig:T_Flow}). Following \citet{imfeld2024transformer}, at the layer entrance of $f'_i$, $\mathbf{T}_\text{in}$ is initialized as the identity matrix $\mathbf{I}$. For residual connections (i.e., Pre-MLP Normalization), $\mathbf{T}_\text{in}$ is set by averaging the $\mathbf{T}_\text{out}$ from both paths (i.e. layer entrance and attention output).

However, vanilla TMF is no longer applicable. We therefore modify the TMF design, these modifications are further ablated in Section~\ref{sec:abl}.

\paragraph{Architectural Advancements.}
Group-Query Attention \citep{ainslie-etal-2023-gqa} and SwiGLU MLP \citep{shazeer2020gluvariantsimprovetransformer} introduce dimensional mismatches that makes vanilla TMF inapplicable. We thus set $\text{T}_{\text{in}} = \mathbf{I}$ for Attention Output and MLP Down. As Group-Query Attention and SwiGLU MLP are widely used in current LLMs, including LLaMA \citep{grattafiori2024llama3herdmodels}, Qwen \citep{yang2025qwen3technicalreport}, and OLMo \citep{olmo20252olmo2furious}, our modified TMF covers a broad range of mainstream architectures.

\paragraph{Non-consecutive Alignment.}
Alignment is applied only to non-consecutive $f'_i$. $\mathbf{T}_{in}$ at layer entrance does not carry meaningful cross-layer information. Incorporating this will dilute the impact of alignment. Thus, we set $\mathbf{T}_{in}=\mathbf{T}_O$ for MLP Gate and Up that after the residual connection.

\subsection{Weight Initialization with OT}
\label{sec:weight_init}

Given the parameters of layers $f_i$ and $f_{i+1}$ and the pre-defined TMF, Algorithm~$1$ demonstrates how to initialize the new layer $f'_i$ via OT. The overall procedure consists of five steps, described below.

\begin{algorithm}[!ht]
\caption{Optimal Transport Depth Up-Scaling}
\label{alg:OT-DUS}
\footnotesize
\begin{algorithmic}[1]

\Require  
$\mathbf{W}_{b}^{(i)} $, $\mathbf{W}_{b}^{(i+1)}$; 
$\mathcal{TMF}=\{\mathbf{T}_{\text{in}}^{(b)}\}_{b=1}^B$ (Table~\ref{fig:T_Flow})

\Ensure 
$\mathbf{W'}_{b}^{(i)}$

\For{base layer $f_i$ $(\frac{n}{2} \le i < n)$}
\For{each module $b$}

\State Initialize
$\mu$ and $\nu$ 
\Comment{OT instantiation}

\State $\mathbf{C}_{kj}=\|\delta(x^{(k)})-\delta(y^{(j)})\|_2$
\Comment{OT instantiation}

\State $\mathbf{T}_{\text{in}} \gets \mathcal{TMF}[b]$
\Comment{Retrieve $\mathbf{T}_{\text{in}}$ from $\mathcal{TMF}$}

\State $\mathbf{W}_{b}^{(i)} \gets
\mathbf{W}_{b}^{(i)} \cdot \mathbf{T}_{\text{in}}$
\Comment{Alignment within layer}

\State $\mathbf{T}_{\text{out}} =
\text{OT}(\mu, \nu, \mathbf{C})$
\Comment{Solve OT problem}

\State $\mathbf{W}_{b}^{(i)} \gets
\mathbf{T}_{\text{out}}^{\top} \cdot \mathbf{W}_{b}^{(i)}$
\Comment{Align across layer}

\State $\mathbf{W'}_{b}^{(i)} \gets\frac{1}{2}(\mathbf{W}_{b}^{(i)} + \mathbf{W}_{b}^{(i+1)})$
\Comment{New Weights}

\EndFor

\State $\mathbf{W'}_{O}^{(i)}, \mathbf{W'}_{\text{Down}}^{(i)} \gets \mathbf{0}$
\Comment{Zero-init}

\EndFor

\end{algorithmic}
\end{algorithm}

\noindent \textbf{Step-1: OT problem instantiation.} 
Based on the definition of OT, we initialize $\mu$ and $\nu$ over $\mathbf{W}_{b}^{(i)}$ and $\mathbf{W}_{b}^{(i+1)}$. Their probability distributions $\boldsymbol{\alpha}$ and $\boldsymbol{\beta}$ are initialized uniformly, treating each neurons equally (cf. line 3).
To measure neuron differences, we instantiate the unit mass $\delta$ using weight-based support \citep{NEURIPS2020_fb269786}, representing each neuron directly by its weight value without auxiliary constraints.
The transport cost $\mathbf{C}_{kj}$ is defined as the Euclidean distance between the $k$-th neuron in $\mathbf{W}_{b}^{(i)}$ and the $j$-th neuron in $\mathbf{W}_{b}^{(i+1)}$ (cf. line 4).

\noindent \textbf{Step-2: Alignment within layer.}  The permutation change caused by aligning $\mathbf{W}_{b-1}^{(i)}$ to $\mathbf{W}_{b-1}^{(i+1)}$ disrupts the original neuron correspondence between $\mathbf{W}_{b-1}^{(i)}$ and $\mathbf{W}_{b}^{(i)}$. Such permutation change information is stored in $\mathbf{T}_\text{in}$ for $\mathbf{W}_{b}^{(i)}$. To restore this, $\mathbf{W}_{b}^{(i)}$ needs to align with $\mathbf{W}_{b-1}^{(i)}$ using $\mathbf{T}_\text{in}$. $\mathbf{T}_\text{in}$ is defined by $\mathcal{TMF}$ for each module, shown in Table~\ref{fig:T_Flow}. After retrieving $\mathbf{T}_\text{in}$ (cf. line 5), the alignment within the layer is performed via $\mathbf{W}_{b}^{(i)} \leftarrow \mathbf{W}_{b}^{(i)} \cdot \mathbf{T}_\text{in}$ (cf. line 6).

\noindent \textbf{Step-3: Alignment across layer} 
We then solve $\text{OT}({\mu},{\nu},\mathbf{C})$ to compute the transport matrix. \citet{imfeld2024transformer} found that the Sinkhorn-Knopp algorithm \cite{10.1137/060659624} is optimal for solving $\text{OT}({\mu},{\nu},\mathbf{C})$ in Transformer fusion. We employ this approach to obtain $\mathbf{T}_\text{out}$ for $\mathbf{W}_{b}^{(i)}$ (cf. line 7). $\mathbf{W}_{b}^{(i)}$ is then aligned with $\mathbf{W}_{b}^{(i+1)}$ using the computed $\mathbf{T}_\text{out}$ via $\mathbf{W}_{b}^{(i)} \leftarrow \mathbf{T}_\text{out}^\top \cdot \mathbf{W}_{b}^{(i)}$ (cf. line 8).

\noindent \textbf{Step-4: Compute new-layer weights.}
$\mathbf{W'}_{b}^{(i)}$ is the average of aligned
$\mathbf{W}_{b}^{(i)}$ and $\mathbf{W}_{b}^{(i+1)}$ (cf. line~9).

\noindent \textbf{Step-5: Zero-Init.}
We set $\mathbf{T}_\text{in}=\mathbf{I}$ for $\mathbf{W}_O$ and $\mathbf{W}_{down}$ in TMF due to architectural advancements (Section~\ref{sec:TMF}), which may cause misalignment problem. Inspired by zero-init \citep{wu-etal-2024-llama}, we set $\mathbf{W}_O = 0$ and $\mathbf{W}_{down} = 0$ (cf. line 11), which naturally resolves this issue while ensuring function preservation, a property crucial for retaining  model performance \citep{wang2024lemon,wu-etal-2024-llama}.

\section{Experimental Setup}

\subsection{Base Model}

Following prior work \citep{wu-etal-2024-llama,kim-etal-2024-solar,yang-etal-2025-lesa}, we use off-the-shelf 32-layer Llama-3.1-8B \citep{grattafiori2024llama3herdmodels} as our \textit{base} model. We further conducted a smaller-scale experiment using off-the-shelf 16-layer Llama-3.2-1B. Following \citet{yang-etal-2025-lesa}, we insert a number of new layers equal to 50\% of the base layers. The \textit{expanded} model sizes are fixed at 11.5B parameters with 48 layers (adding 16 layers) and 1.72B with 24 layers (adding 8 layers) for all depth up-scaling methods. 

\subsection{Baselines}
We experiment with state-of-the-art depth up-scaling methods, as shown in Figure~\ref{fig:main}. Hyper-parameter and implementation details are provided in Appendix \ref{App:hyper} and \ref{App:imp}.

\noindent \textbf{Base.} We continue pre-training the \textit{base} model without expansion. All layers are trained.

\noindent \textbf{SOLAR.} This method copies the bottom and top $m$ layers from $\mathcal{M}$ to form $\mathcal{M'}$. We choose $m=24$ and $m=12$ for 11.5B and 1.72B \textit{expanded} models. All layers are trained in line with \citet{kim-etal-2024-solar}.

\noindent \textbf{LLaMA PRO.} It divides $\mathcal{M}$ into $g$ groups of $m$ layers. $p$ new layers are created by copying the top-$p$ base layers and inserted on top of each group with zero-init. We use $g=16$ (11.5B \textit{expanded} models) and $g=8$ (1.72B \textit{expanded} models); $m=2$ and $p=1$ are used throughout. Only $f'$ are trained following \citet{wu-etal-2024-llama}.

\noindent \textbf{LESA.} This approach uses an auxiliary network to initialize $f_i'$ given $f_i$ and $f_{i+1}$. LESA inserts $f_i'$ in the top half of $\mathcal{M}$. We insert new layers between $f_{16}$ and $f_{32}$ (11.5B \textit{expanded} models), and between $f_{8}$ to $f_{16}$ (1.72B \textit{expanded} models). Only $f'$ are trained as in \citet{yang-etal-2025-lesa}.

\subsection{Training Data}
For Continual Pre-Training (CPT), we opt using data of same size as in \citet{yang-etal-2025-lesa}, published after the base model's knowledge cut-off. We sample 1.5B tokens from the CC-MAIN-2024-51 subset of FineWeb-Edu \citep{NEURIPS2024_370df50c}. For supervised fine-tuning (SFT), we choose Alpaca GPT4 \citep{peng2023instructiontuninggpt4} and update the whole model following \citet{yang-etal-2025-lesa}.

\subsection{Evaluation} 

Following previous studies \citep{wu-etal-2024-llama,yang-etal-2025-lesa}, we conduct experiments focusing on general knowledge-related tasks. We further conduct extensive experiments on specialized domains: biomedical \citep{10.1093/bioinformatics/btz682, 10.1145/3458754,10.1093/bib/bbac409, singhal2023expertlevelmedicalquestionanswering} and legal \citep{chalkidis-etal-2019-neural, 10.1145/3462757.3466088,  NEURIPS2022_bc218a0c, niklaus-etal-2024-multilegalpile} as they are widely explored within LLMs. Following common practice in LLM evaluation \citep{arabelly-etal-2025-improving,li2025active,cao-etal-2026-moa}, we report results averaged over three random seeds.

\noindent \textbf{General.}
 We include ARC-Easy \citep{clark2018thinksolvedquestionanswering}, LogiQA \citep{ijcai2020p501}, Winogrande \citep{10.1145/3474381} for \textit{Reasoning}; CSQA \citep{talmor-etal-2019-commonsenseqa}, BoolQ \citep{clark-etal-2019-boolq}, PIQA \citep{Bisk2020} for \textit{Commonsense and Knowledge};  MMLU \citep{hendrycks2021measuring} for \textit{Examination}; and WikiText \citep{merity2017pointer} for \textit{Language Modeling}.

\noindent \textbf{Biomedical.}
Following previous work \citep{williams2025compressinglanguagemodelsspecialized}, we include MultiMedQA benchmark \citep{singhal2022largelanguagemodelsencode}, specifically the PubMedQA \citep{jin-etal-2019-pubmedqa}, MedQA \citep{app11146421}, MedMCQA \citep{pmlr-v174-pal22a} tasks, and relevant subsets from MMLU \citep{hendrycks2021measuring} (anatomy, clinical knowledge, college medicine, medical genetics, professional medicine, college biology).

\noindent \textbf{Legal.}
We follow \citet{williams2025compressinglanguagemodelsspecialized} in using CaseHOLD \citep{10.1145/3462757.3466088} and ECtHR (Task A) \citep{chalkidis-etal-2019-neural} from LexGLUE \citep{chalkidis-etal-2022-lexglue} and Legal-MMLU, covering jurisprudence, professional law, and international law \citep{hendrycks2021measuring}.

\begin{table*}[!ht]
\begin{center}
\footnotesize
\setlength{\tabcolsep}{1mm}
\resizebox{0.95\textwidth}{!}{
\begin{tabular}{c|lccccccccc} 
\toprule
\multicolumn{2}{c}{} & \multicolumn{1}{c}{\textbf{Perplexity\,$\downarrow$}}
& \multicolumn{7}{c}{\textbf{Zero-shot Performance\,$\uparrow$}} \\ 
\cmidrule(l{2pt}r{2pt}){3-3} \cmidrule(l{2pt}r{2pt}){4-10} 
\multicolumn{1}{c}{} & \multicolumn{1}{c}{\textbf{Methods}} 
& \textbf{Wiki-PPL} & \textbf{ARC} & \textbf{LogiQA} & \textbf{Wino} & \textbf{CSQA} & \textbf{BoolQ} & \textbf{PIQA} & \textbf{MMLU} & \textbf{Average} \\ 
\midrule

\multirow{6}{*}{\rotatebox[origin=c]{90}{\textbf{CPT}}}

& Base-8B          
& $8.35_{0.00}$ 
& $79.91_{0.05}$ 
& $26.68_{0.18}$ 
& $72.19_{0.30}$ 
& $65.19_{0.08}$ 
& $81.85_{0.04}$ 
& $79.04_{0.26}$ 
& $58.86_{0.22}$ 
& $66.25_{0.04}$ \\

& SOLAR-11.5B           
& $9.90_{0.00}$ 
& $80.02_{0.15}$ 
& $26.88_{0.00}$ 
& $71.80_{0.24}$ 
& $57.36_{0.05}$ 
& $80.76_{0.05}$ 
& $78.51_{0.05}$ 
& $54.32_{0.05}$ 
& $64.24_{0.05}$ \\

& LLaMA PRO-11.5B       
& $7.81_{0.00}$ 
& $81.45_{0.15}$ 
& $\mathbf{29.39}_{\mathbf{0.09}}$ 
& $73.93_{0.24}$ 
& $70.71_{0.21}$ 
& $81.71_{0.06}$ 
& $79.89_{0.08}$ 
& $62.59_{0.03}$ 
& $68.52_{0.04}$ \\

& LESA-11.5B            
& \underline{$7.73_{0.00}$} 
& \underline{$81.92_{0.15}$} 
& \underline{$28.06_{0.09}$} 
& \underline{$74.09_{0.12}$} 
& $\mathbf{72.32}_{\mathbf{0.14}}$ 
& \underline{$81.89_{0.04}$} 
& \underline{$80.36_{0.06}$} 
& \underline{$62.61_{0.03}$} 
& \underline{$68.75_{0.03}$} \\

& OT-DUS-11.5B (Ours)        
& $\mathbf{7.73}_{\mathbf{0.00}}$ 
& $\mathbf{82.06}_{\mathbf{0.06}}$ 
& $27.34_{0.00}$ 
& $\mathbf{74.77}_{\mathbf{0.20}}$ 
& \underline{$71.96_{0.33}$} 
& $\mathbf{82.18}_{\mathbf{0.07}}$ 
& $\mathbf{80.56}_{\mathbf{0.22}}$ 
& $\mathbf{62.99}_{\mathbf{0.02}}$ 
& $\mathbf{68.84}_{\mathbf{0.04}}$ \\

\midrule

\multirow{6}{*}{\rotatebox[origin=c]{90}{\textbf{SFT}}}

& Base-8B           
& $8.32_{0.00}$ 
& $81.07_{0.09}$ 
& $24.73_{0.15}$ 
& $72.17_{0.28}$ 
& $68.25_{0.25}$ 
& $82.06_{0.08}$ 
& $79.85_{0.12}$ 
& $59.38_{0.19}$ 
& $66.79_{0.05}$ \\

& SOLAR-11.5B          
& $9.68_{0.00}$ 
& $80.81_{0.11}$ 
& $25.29_{0.09}$ 
& $71.19_{0.24}$ 
& $60.99_{0.21}$ 
& $81.34_{0.13}$ 
& $79.14_{0.03}$ 
& $55.21_{0.17}$ 
& $64.85_{0.07}$ \\

& LLaMA PRO-11.5B      
& $7.81_{0.00}$ 
& $83.30_{0.05}$ 
& $\mathbf{27.39}_{\mathbf{0.18}}$ 
& $73.93_{0.16}$ 
& $72.07_{0.16}$ 
& $82.26_{0.03}$ 
& \underline{$80.81_{0.08}$} 
& $62.48_{0.14}$ 
& $68.89_{0.03}$ \\

& LESA-11.5B          
& $\mathbf{7.72}_{\mathbf{0.00}}$ 
& $\mathbf{83.84}_{\mathbf{0.04}}$ 
& \underline{$26.83_{0.24}$} 
& \underline{$75.61_{0.36}$} 
& \underline{$72.97_{0.29}$} 
& \underline{$82.72_{0.24}$} 
& $80.52_{0.20}$ 
& \underline{$63.82_{0.22}$} 
& \underline{$69.48_{0.05}$} \\

& OT-DUS-11.5B (Ours)       
& \underline{$7.73_{0.00}$} 
& \underline{$83.80_{0.04}$} 
& $26.73_{0.00}$ 
& $\mathbf{75.88}_{\mathbf{0.44}}$ 
& $\mathbf{73.00}_{\mathbf{0.09}}$ 
& $\mathbf{83.31}_{\mathbf{0.04}}$ 
& $\mathbf{80.89}_{\mathbf{0.11}}$ 
& $\mathbf{63.99}_{\mathbf{0.15}}$ 
& $\mathbf{69.65}_{\mathbf{0.07}}$ \\

\midrule
\midrule

\multirow{5}{*}{\rotatebox[origin=c]{90}{\textbf{CPT}}}

& Base-1B         
& $13.68_{0.00}$ 
& \underline{$68.64_{0.25}$} 
& \underline{$21.56_{0.18}$} 
& $58.43_{0.09}$ 
& $24.57_{0.16}$ 
& $62.69_{0.32}$ 
& $75.03_{0.10}$ 
& $29.27_{0.37}$ 
& $48.60_{0.14}$ \\

& SOLAR-1.72B           
& $13.87_{0.00}$ 
& $\mathbf{69.00}_{\mathbf{0.10}}$ 
& $21.35_{0.15}$ 
& $59.54_{0.12}$ 
& $21.54_{0.71}$ 
& $61.12_{0.09}$ 
& $74.88_{0.27}$ 
& $28.70_{0.14}$ 
& $48.02_{0.15}$ \\

& LLaMA PRO-1.72B       
& $12.43_{0.00}$ 
& $67.43_{0.17}$ 
& $20.99_{0.09}$ 
& $\mathbf{61.48}_{\mathbf{0.47}}$ 
& $34.23_{0.25}$ 
& \underline{$62.83_{0.10}$} 
& $\mathbf{75.57}_{\mathbf{0.05}}$ 
& $31.81_{0.10}$ 
& $50.62_{0.09}$ \\

& LESA-1.72B            
& \underline{$12.28_{0.00}$} 
& $66.70_{0.02}$ 
& $21.45_{0.23}$ 
& $59.64_{0.87}$ 
& \underline{$40.89_{0.12}$} 
& $\mathbf{63.72}_{\mathbf{0.09}}$ 
& $74.74_{0.08}$ 
& $\mathbf{33.27}_{\mathbf{0.21}}$ 
& \underline{$51.49_{0.11}$} \\

& OT-DUS-1.72B  (Ours)      
& $\mathbf{12.19}_{\mathbf{0.00}}$ 
& $67.10_{0.09}$ 
& $\mathbf{22.84}_{\mathbf{0.23}}$ 
& \underline{$60.06_{0.61}$} 
& $\mathbf{42.97}_{\mathbf{0.37}}$ 
& $62.75_{0.08}$ 
& \underline{$75.17_{0.14}$} 
& \underline{$32.88_{0.13}$} 
& $\mathbf{51.97}_{\mathbf{0.07}}$ \\

\midrule

\multirow{5}{*}{\rotatebox[origin=c]{90}{\textbf{SFT}}}

& Base-1B    
& $13.57_{0.00}$ 
& \underline{$69.85_{0.11}$} 
& \underline{$21.97_{0.41}$} 
& $59.35_{0.44}$ 
& $25.83_{0.45}$ 
& $62.94_{0.18}$ 
& $75.72_{0.13}$ 
& $30.17_{0.24}$ 
& $49.40_{0.12}$ \\

& SOLAR-1.72B        
& $13.68_{0.00}$ 
& $\mathbf{70.41}_{\mathbf{0.04}}$ 
& $\mathbf{22.22}_{\mathbf{0.09}}$ 
& $59.11_{0.16}$ 
& $25.25_{0.54}$ 
& $60.88_{0.09}$ 
& \underline{$75.86_{0.03}$} 
& $29.79_{0.35}$ 
& $49.08_{0.13}$ \\

& LLaMA PRO-1.72B   
& $\mathbf{12.36}_{\mathbf{0.00}}$ 
& $68.08_{0.06}$ 
& $21.10_{0.23}$ 
& $\mathbf{60.62}_{\mathbf{0.32}}$ 
& $38.19_{0.42}$ 
& $64.09_{0.06}$ 
& $76.03_{0.21}$ 
& $30.99_{0.22}$ 
& $51.30_{0.07}$ \\

& LESA-1.72B        
& $12.54_{0.00}$ 
& $67.86_{0.11}$ 
& $20.84_{0.09}$ 
& $60.25_{0.60}$ 
& \underline{$43.73_{0.25}$} 
& \underline{$65.05_{0.20}$} 
& $75.81_{0.06}$ 
& $\mathbf{34.98}_{\mathbf{0.44}}$ 
& \underline{$52.65_{0.16}$} \\

& OT-DUS-1.72B (Ours)    
& \underline{$12.46_{0.00}$} 
& $68.28_{0.13}$ 
& $21.15_{0.32}$ 
& \underline{$60.30_{0.27}$} 
& $\mathbf{44.69}_{\mathbf{0.19}}$ 
& $\mathbf{65.76}_{\mathbf{0.07}}$ 
& $\mathbf{75.90}_{\mathbf{0.10}}$ 
& \underline{$33.22_{0.11}$} 
& $\mathbf{52.76}_{\mathbf{0.04}}$ \\

\bottomrule
\end{tabular}
}
\end{center}
\caption{CPT on 1.5B tokens and SFT (after CPT) performance of 11.5B and 1.72B \textit{expanded} models.}
\label{8B-Expanded}
\end{table*}

\section{Experimental Results}
\label{sec:performance}

\subsection{General Performance}\label{sec:KT}
Table~\ref{8B-Expanded} (Top) presents the CPT and SFT results of 11.5B \textit{expanded} models. For CPT, we observe that OT-DUS achieves top performance on six out of eight benchmarks, specifically Wiki-PPL (7.73), ARC (82.06) ,Winogrande (74.77), BoolQ (82.18), PIQA (80.56), MMLU (62.99). This strong performance across various downstream tasks, yielding the highest average score (68.84), highlights the effectiveness. We note that the strong performance continues in SFT. It achieves top performance on Winogrande, CSQA, BoolQ, PIQA, MMLU and second performance on Wiki-PPL, and ARC, with the top average score (69.65). 

To analyze performance with different training budgets, we save five checkpoints while training the 11.5B \textit{expanded} models (every 20\% training steps), shown in Appendix~\ref{App:performance across checkpoints}. We observe that OT-DUS consistently achieves top performance on at least five out of eight benchmarks across all checkpoints.

\subsection{Performance at Smaller Scales}
Table~\ref{8B-Expanded} (Bottom) presents the CPT and SFT results of the 1.72B \textit{expanded} models. For CPT, OT-DUS achieves the highest average score (51.97), ranking first on Wiki-PPL (12.19), LogiQA (22.84), and CSQA (42.97), while ranking second on Winogrande, PIQA, and MMLU. For SFT, OT-DUS again achieves the highest average score, ranking first on CSQA, BoolQ, and PIQA, and second on Wiki-PPL, Winogrande, and MMLU.

Compared with LESA, the strongest baseline on average, OT-DUS achieves top-2 results on more benchmarks in both CPT and SFT settings (Table~\ref{8B-Expanded}). In CPT, OT-DUS achieves two best and three second-best results at 1.72B, and five best and one second-best results at 11.5B, compared with LESA's two best and one second-best, and one best and six second-best results, respectively. In SFT, OT-DUS achieves three best and two second-best results at 1.72B, and five best and one second-best results at 11.5B, compared with LESA's one best and two second-best, and one best and five second-best results.  These results demonstrate the consistent effectiveness of OT-DUS across model scales, training stages, and benchmarks.

\subsection{Up-scaling Stability at Larger Scales}
\label{sec:up-scale stability}

We follow previous work \citep{yano-etal-2025-step,yang-etal-2025-lesa} by reporting perplexity without any training to evaluate up-scaling stability on larger models. Appendix~\ref{App:Up-scale stability} presents the perplexity at different model scales. We observe that both LLaMA-Pro and OT-DUS match the base model's perplexity regardless of model parameters due to zero-init, demonstrating maximum expansion stability compared to other baselines. Surprisingly, we find that LESA's perplexity sharply increases when applied to Llama-3.2-1B (871.50). We hypothesize this is because smaller models have fewer layers. This leads to less training data for the auxiliary network, consequently causing it to underfit.

\subsection{Domain Specific Performance}

Table~\ref{domain-specific} presents the performance on biomedical and legal domains. We observe OT-DUS achieves the best overall performance. In biomedical tasks, OT-DUS wins two out of four tasks (i.e. MedMCQA and MedQA), while offering the second best performance on PubMedQA. Strong performance is also observed in the legal domain, where OT-DUS wins two out three legal tasks and achieves second best on the remaining one. 

\begin{table}[!ht]
\begin{center}
\footnotesize
\setlength{\tabcolsep}{1mm}

\resizebox{\columnwidth}{!}{
\begin{tabular}{c|lcccc}
\toprule

\multicolumn{1}{c}{}
& \textbf{Methods}
& \textbf{MedMCQA}
& \textbf{MedQA}
& \textbf{Bio-MMLU}
& \textbf{PubMedQA} \\
\midrule

\multirow{5}{*}{\rotatebox{90}{\textbf{Biomedical}}}
& Base       
& $48.92_{0.28}$ 
& $53.26_{0.00}$ 
& $68.35_{0.10}$ 
& \underline{$76.33_{0.12}$} \\

& SOLAR      
& $45.22_{0.13}$ 
& $47.24_{0.05}$ 
& $57.41_{0.14}$ 
& $76.20_{0.00}$ \\

& LLaMA PRO  
& $54.47_{0.22}$ 
& $59.31_{0.14}$ 
& \underline{$71.16_{0.27}$} 
& $\mathbf{77.67}_{\mathbf{0.12}}$ \\

& LESA       
& \underline{$56.73_{0.19}$} 
& \underline{$60.07_{0.18}$} 
& $\mathbf{71.40}_{\mathbf{0.36}}$ 
& $76.00_{0.00}$ \\

& OT-DUS   
& $\mathbf{56.79}_{\mathbf{0.16}}$ 
& $\mathbf{60.36}_{\mathbf{0.09}}$ 
& $71.30_{0.29}$ 
& \underline{$76.33_{0.12}$} \\

\midrule

\multicolumn{1}{c}{}
& \textbf{Methods}
& \multicolumn{1}{c}{\textbf{Legal-MMLU}}
& \multicolumn{2}{c}{\textbf{CaseHOLD}}
& \multicolumn{1}{c}{\textbf{ECtHR}} \\
\midrule

\multirow{5}{*}{\rotatebox{90}{\textbf{Legal}}}
& Base       
& \multicolumn{1}{c}{$64.48_{0.01}$}
& \multicolumn{2}{c}{$47.42_{0.06}$}
& \multicolumn{1}{c}{$56.06_{0.12}$} \\

& SOLAR      
& \multicolumn{1}{c}{$62.62_{0.15}$}
& \multicolumn{2}{c}{$41.02_{0.13}$}
& \multicolumn{1}{c}{$38.90_{0.26}$} \\

& LLaMA PRO  
& \multicolumn{1}{c}{$\mathbf{68.58}_{\mathbf{0.15}}$}
& \multicolumn{2}{c}{$48.31_{0.17}$}
& \multicolumn{1}{c}{$60.40_{0.12}$} \\

& LESA       
& \multicolumn{1}{c}{$67.14_{0.11}$}
& \multicolumn{2}{c}{\underline{$51.06_{0.06}$}}
& \multicolumn{1}{c}{\underline{$61.10_{0.19}$}} \\

& OT-DUS   
& \multicolumn{1}{c}{\underline{$68.14_{0.18}$}}
& \multicolumn{2}{c}{$\mathbf{51.91}_{\mathbf{0.09}}$}
& \multicolumn{1}{c}{$\mathbf{61.22}_{\mathbf{0.07}}$} \\

\bottomrule
\end{tabular}
}

\end{center}
\caption{Domain-specific CPT performance of 11.5B \textit{expanded} models.}
\label{domain-specific}
\end{table}

\section{Experimental Analysis}

\subsection{Interpolation Positions}
\label{sec:IP}

We extensively investigate the best interpolation positions of OT-DUS. We evaluate the following strategies: inserting in the bottom half (Btm), in the middle portion (Mid), in the top half (Top), and at the top and bottom quarters (T\&B). The layer index ranges are defined as follows:
{
\begin{equation*}
\footnotesize
\begin{aligned} 
\mathcal{M}'(x; \theta') \circ =
\begin{cases}
f'_i \circ f_i,  \:\: i \leq \frac{n}{2}  & \text{if Btm}  \\
f'_i \circ f_i,  \:\: \frac{n}{4} < i \leq \frac{3n}{4}& \text{if Mid}  \\
f'_i \circ f_i,  \:\: \frac{n}{2} \leq i < n& \text{if Top}  \\
f'_i \circ f_i,  \:\: i \leq  \frac{n}{4} \:\:\text{or}\:\: \frac{3n}{4} \leq i < n& \text{if T\&B}  \\
\end{cases}
\end{aligned}
\end{equation*}
}

Table~\ref{opt_deus_variants} illustrates the performance of different interpolation strategies. We observe that OT-DUS-Top is the best performing strategy, overall. It achieves the top performance on five out of eight benchmarks, while obtaining the second best performance on two benchmarks. The performance difference between interpolation strategies is consistent with previous work, where inserting new layers into the top part offers additional performance gains \citep{yang-etal-2025-lesa}. This further supports previous findings showing that bottom layers in Transformers are more critical \citep{jawahar-etal-2019-bert}, while top layers are less sensitive to modification \citep{men-etal-2025-shortgpt}.

\begin{table}[ht]
\centering
\footnotesize
\resizebox{0.7\linewidth}{!}{
\begin{tabular}{lc}
\toprule
\textbf{Interpolation Strategy} & \textbf{Average} \\
\midrule
OT-DUS-Btm  & $68.39_{0.12}$\\
OT-DUS-Mid  & $68.28_{0.01}$ \\
OT-DUS-Top  & $\mathbf{68.84}_\mathbf{0.04}$ \\
OT-DUS-T\&B & \underline{$68.44_{0.04}$}\\
\bottomrule
\end{tabular}}
\caption{Average performance of OT-DUS-11.5B using different interpolation strategies. Full results are presented in Table \ref{tab:opt_deus_variants} (Appendix \ref{App:Interpolation Strategy}).}
\label{opt_deus_variants}
\end{table}

\subsection{Training Efficiency}
\label{TE}

Previous work ignores the impact of interpolation strategy regarding training efficiency \citep{wu-etal-2024-llama,yang-etal-2025-lesa}. Table~\ref{Training Time} shows that progressive depth up-scaling methods considerably outperform SOLAR (22:54:11) in training efficiency. We observe a strong correlation between interpolation positions and efficiency: top-half insertions, exemplified by OT-DUS-Top (12:52:04) and LESA (12:54:07), are notably faster. Conversely, strategies inserting layers in the bottom half, such as OT-DUS-Btm (14:56:00) and LLaMA PRO (14:58:34), require longer training time. This pattern holds regardless of the initialization method. The efficiency differences mainly arise from higher backpropagation costs when updating layers inserted at lower positions in the model.

\begin{table}[!ht]
\centering
\footnotesize
\setlength{\tabcolsep}{1mm}
\resizebox{0.9\linewidth}{!}{
\begin{tabular}{@{}lcl@{}}
\toprule
\textbf{Methods} & \textbf{Trainable}  & \textbf{Training Time} \\
\midrule
SOLAR-11.5B         & 11B  & 22:54:11 (+78.0\%) \\
LLaMA PRO-11.5B     & 4B   & 14:58:34 (+16.4\%) \\
LESA-11.5B          & 4B   & 12:54:07 (+0.3\%) \\
\midrule
OT-DUS-Btm-11.5B  & 4B   & 14:56:00 (+16.1\%) \\
OT-DUS-Mid-11.5B  & 4B   & 13:53:14 (+7.9\%) \\
OT-DUS-Top-11.5B  & 4B   & 12:52:04 \\
OT-DUS-T\&B-11.5B & 4B   & 14:45:38 (+14.7\%) \\
\bottomrule
\end{tabular}
}
\caption{Training time for 11.5B models.}
\label{Training Time}
\end{table}

OT-DUS incurs additional computation for solving the OT problem. Compared with baselines that also require extra computation (Table~\ref{Extra Time}, Appendix~\ref{App:Additional Computation Cost}), it introduces less overhead and achieves the best overall efficiency among the baselines.

\subsection{Ablation Studies}
\label{sec:abl}

\begin{table}[ht]
\begin{center}
\footnotesize
\renewcommand{\arraystretch}{1.25}
\resizebox{\columnwidth}{!}{
\begin{tabular}{c|lc}
\toprule
\multicolumn{1}{c}{} 
& \multicolumn{1}{l}{\textbf{Ablation Variants}}
& \textbf{Average} \\
\midrule

\multirow{2}{*}{\rotatebox[origin=c]{90}{\textbf{TMF}}}
& Attention Output $\mathbf{T}_\text{in}=\mathbf{T}_Q$-1.72B  & \underline{$51.96_{0.03}$} \\
&  MLP Gate and Up $\mathbf{T}_\text{in}=\frac{1}{2}(\mathbf{T}_O + \mathbf{I})$)-1.72B  & $51.79_{0.02}$ \\

\midrule

\multirow{3}{*}{\rotatebox[origin=c]{90}{\textbf{OT}}}
& Average-1.72B                    & $51.59_{0.61}$ \\
& Copy-1.72B                       & $50.13_{0.01}$ \\
& Random-1.72B                     & $51.39_{0.13}$ \\

\midrule

\multirow{3}{*}{\rotatebox[origin=c]{90}{\textbf{Zero-Init}}}
& Average + Zero-Init-1.72B        & $51.20_{0.61}$ \\
& Copy + Zero-Init-1.72B           & $51.67_{0.00}$ \\
& Random + Zero-Init-1.72B         & $51.58_{0.00}$ \\

\midrule

& OT-DUS-1.72B (Ours)          & $\mathbf{51.97}_\mathbf{0.07}$ \\

\midrule
\midrule

\multirow{2}{*}{\rotatebox[origin=c]{90}{\textbf{TMF}}}
& Attention Output $\mathbf{T}_\text{in}=\mathbf{T}_Q$-11.5B & $68.71_{0.02}$ \\
&  MLP Gate and Up $\mathbf{T}_\text{in}=\frac{1}{2}(\mathbf{T}_O + \mathbf{I})$)-11.5B  & $68.73_{0.01}$ \\

\midrule

\multirow{3}{*}{\rotatebox[origin=c]{90}{\textbf{OT}}}
& Average-11.5B                    &  $68.44_{0.04}$ \\
& Copy-11.5B                       &  $68.59_{0.01}$ \\
& Random-11.5B                     &  $67.11_{0.05}$ \\

\midrule

\multirow{3}{*}{\rotatebox[origin=c]{90}{\textbf{Zero-Init}}}
& Average + Zero-Init-11.5B        &  $68.62_{0.03}$ \\
& Copy + Zero-Init-11.5B           & \underline{ $68.82_{0.06}$} \\
& Random + Zero-Init-11.5B         &  $68.43_{0.06}$ \\

\midrule

& OT-DUS-11.5B (Ours)          & $\mathbf{68.84}_\mathbf{0.04}$ \\

\bottomrule
\end{tabular}
}
\end{center}
\caption{Average performance of ablation variants. Full Results are presented at Table \ref{tab:ablation_full} (Appendix \ref{App:ablations}).}
\label{tab:ablation_design}
\end{table}

\paragraph{Ablation on TMF design.} We modify the vanilla TMF to account for architectural advancements and non-consecutive alignment (Section~\ref{sec:TMF}). We further compare our design with alternative variants for architectural advancements (Output: $\mathbf{T}_\text{in}=\mathbf{T}_Q$) and non-consecutive alignment (Gate/Up: $\mathbf{T}_\text{in}=\frac{1}{2}(\mathbf{T}_O + \mathbf{I})$). Table~\ref{tab:ablation_design} shows that OT-DUS achieves the best overall performance (Avg: $51.97$), validating our TMF design.

\paragraph{Ablation on OT alignment.} To validate OT (Section~\ref{sec:motivation}), we compare it with direct copy, average, and random (Table~\ref{tab:ablation_design}). Compared with the second-best variant, Copy + Zero-Init, OT-DUS outperforms it on 4 of 7 benchmarks at 11.5B and 6 of 7 benchmarks at 1.72B (Table~\ref{tab:ablation_full}), demonstrating consistent gains across model scales and tasks. These results demonstrate that OT alignment provides a more meaningful initialization that mitigates neuron permutation mismatch.

\paragraph{Ablation on Zero-Init.} OT-DUS adopts zero-init (Section~\ref{sec:weight_init}). We further compare it with non-OT variants using zero-init (Table~\ref{tab:ablation_design}). OT-DUS consistently outperforms these variants, while zero-init improves performance for non-OT variants. Together with the observed up-scaling stability (Section~\ref{sec:up-scale stability}), these results support the use of zero-init.

\paragraph{Ablation on Model families.} We further conduct experiments using Qwen-3-8B \citep{yang2025qwen3technicalreport} as \textit{base} model. Table \ref{Qwen-8B} shows that OT-DUS achieves the best overall performance, obtaining the top performance on five out of eight benchmarks. This further validates the generalization of OT-DUS across model families.

\begin{table}[ht]
\centering
\footnotesize
\resizebox{0.6\columnwidth}{!}{
\begin{tabular}{lc}
\toprule
\textbf{Methods} & \textbf{Average} \\
\midrule
Base-8B
& $73.59_{0.01}$ \\

SOLAR-11.5B
& $72.36_{0.00}$ \\

LLaMA PRO-11.5B
& \underline{$73.80_{0.04}$} \\

LESA-11.5B
& $73.76_{0.17}$ \\

OT-DUS-11.5B (Ours)
& $\mathbf{74.47}_{\mathbf{0.06}}$ \\
\bottomrule
\end{tabular}}
\caption{Average Performance of \textit{expanded} models using Qwen3-8B as \textit{base} model. Full results are presented in Table \ref{8B_Qwen_Full} (Appendix \ref{App:ablations}).}
\label{Qwen-8B}
\end{table}

\subsection{Analysis of Neuron Functionality}
To investigate neuron functionality mapping, we analyze the transport matrix between each new layer and its adjacent base layers. Following Algorithm~$1$ (Section~\ref{sec:weight_init}), we instantiate the OT problem and solve it with the same hyperparameters, then compute the entropy \citep{6773024} of each transport matrix (Table~\ref{Entropy}). The transport matrix reflects neuron-level functional correspondence, while entropy measures how diffuse the mapping is. Higher entropy indicates a smoother mapping \citep{NIPS2013_af21d0c9}. SOLAR and LESA are excluded since SOLAR inserts continuous new layers and LESA does not directly use
base layers.

\begin{table}[!ht]
\centering
\footnotesize
\resizebox{\columnwidth}{!}{
\setlength{\tabcolsep}{1mm} 
\begin{tabular}{lcccccccc} 
\toprule
& \multicolumn{4}{c}{\textbf{Attention}}
& \multicolumn{3}{c}{\textbf{MLP}}
&  \\
\cmidrule(lr){2-5}
\cmidrule(lr){6-8}

\textbf{Methods} 
& \textbf{Query} 
& \textbf{Key} 
& \textbf{Value} 
& \textbf{Output} 
& \textbf{Gate} 
& \textbf{Up} 
& \textbf{Down} 
& \textbf{Average} \\
\midrule
LLaMA PRO 
 & 12.23 
 & 9.99 
 & 10.29 
 & 16.64 
 & 14.17 
 & 14.20 
 & 16.63 
 & 13.45  \\
Average
 & 11.69 
 & 7.32 
 & 10.90 
 & 15.60 
 & 17.06 
 & 18.71 
 & 8.56 
 & 12.84  \\
OT-DUS
 & 13.25 
 & 10.32 
 & 11.57 
 & 16.64 
 & 16.78 
 & 18.36 
 & 16.64 
 & 14.79  \\
\bottomrule
\end{tabular}  
}
\caption{Layer-averaged Entropy of the transport matrix between base and new layers in the 11.5B \textit{expanded} models. Normalization components are excluded as they do not involve cross-neuron mixing.} 
\label{Entropy} 
\end{table}

We observe a clear correlation between higher transport matrix entropy and stronger performance. OT-DUS achieves the highest entropy (14.79) and the best performance (68.84), followed by LLaMA PRO (13.45/68.52) and Average (12.84/68.44). Higher entropy indicates that information from each neuron is distributed across multiple neurons in the next layer, forming representations through a richer mixture of input features and a more diverse representational space \citep{e19090474,8998186}, which leads to better performance.

Average yields low entropy for the Key Projection (7.32 vs. 9.99/10.32) and Down Projection (8.56 vs. 16.63/16.64), suggesting these modules are more sensitive to averaging and thus suffer greater performance degradation. Low entropy indicates that functionality from old neurons is concentrated on fewer neurons in the new layer, causing a small subset to dominate the module output. This observation aligns with prior works on the low-rank bottleneck in Query and Key  \citep{pmlr-v119-bhojanapalli20a} and parameter redundancy in Down \citep{pires-etal-2023-one, wei2024investigatinglowranktrainingtransformer}.

\section{Conclusion}

We introduced OT-DUS, a depth up-scaling approach that aligns neurons within and across layers via OT to mitigate neuron permutation mismatches. Results demonstrate that OT-DUS achieves better performance in both general and specialized domains with improved efficiency compared to existing approaches across different model sizes, model families, and training stages. Our systematic analysis of interpolation positions reveals that inserting new layers closer to the top leads to higher training efficiency with improved performance. Our neuron functionality analysis further reveals a correlation between strong performance and high transport matrix entropy when initializing new layers.

\section*{Limitations}
Following recent work \citep{kim-etal-2024-solar,wu-etal-2024-llama,yang-etal-2025-lesa}, we focus on up-scaling already pre-trained LLMs rather than up-scaling during pre-training as in prior work \citep{pmlr-v162-shen22f,wang2023learning,wang2024lemon}. OT-DUS mitigates neuron permutation mismatch by aligning neurons with similar functionalities across adjacent base layers, which implicitly assumes that base layers have already learned relatively stable functionalities. This assumption may not hold during intermediate stages of pre-training, when neurons are still learning new features. Moreover, applying up-scaling during pre-training would require tremendous training budgets, which are beyond our current computational resources.

\section*{Acknowledgments}
We thank Atsuki Yamaguchi for his valuable feedback. We acknowledge IT Services at the University of Sheffield and Bristol Centre for Supercomputing for the provision of HPC services.

\bibliography{custom}

@inproceedings{NEURIPS2020_1457c0d6,
 author = {Brown, Tom and Mann, Benjamin and Ryder, Nick and Subbiah, Melanie and Kaplan, Jared D and Dhariwal, Prafulla and Neelakantan, Arvind and Shyam, Pranav and Sastry, Girish and Askell, Amanda and Agarwal, Sandhini and Herbert-Voss, Ariel and Krueger, Gretchen and Henighan, Tom and Child, Rewon and Ramesh, Aditya and Ziegler, Daniel and Wu, Jeffrey and Winter, Clemens and Hesse, Chris and Chen, Mark and Sigler, Eric and Litwin, Mateusz and Gray, Scott and Chess, Benjamin and Clark, Jack and Berner, Christopher and McCandlish, Sam and Radford, Alec and Sutskever, Ilya and Amodei, Dario},
 booktitle = {Advances in Neural Information Processing Systems},
 editor = {H. Larochelle and M. Ranzato and R. Hadsell and M.F. Balcan and H. Lin},
 pages = {1877--1901},
 publisher = {Curran Associates, Inc.},
 title = {Language {M}odels are {F}ew-{S}hot {L}earners},
 url = {https://proceedings.neurips.cc/paper_files/paper/2020/file/1457c0d6bfcb4967418bfb8ac142f64a-Paper.pdf},
 volume = {33},
 year = {2020}
}

@misc{kaplan2020scalinglawsneurallanguage,
      title={Scaling {L}aws for {N}eural {L}anguage {M}odels}, 
      author={Jared Kaplan and Sam McCandlish and Tom Henighan and Tom B. Brown and Benjamin Chess and Rewon Child and Scott Gray and Alec Radford and Jeffrey Wu and Dario Amodei},
      year={2020},
      eprint={2001.08361},
      archivePrefix={arXiv},
      primaryClass={cs.LG},
      url={https://arxiv.org/abs/2001.08361}, 
}

@article{
wei2022emergent,
title={{E}mergent {A}bilities of {L}arge {L}anguage {M}odels},
author={Jason Wei and Yi Tay and Rishi Bommasani and Colin Raffel and Barret Zoph and Sebastian Borgeaud and Dani Yogatama and Maarten Bosma and Denny Zhou and Donald Metzler and Ed H. Chi and Tatsunori Hashimoto and Oriol Vinyals and Percy Liang and Jeff Dean and William Fedus},
journal={Transactions on Machine Learning Research},
issn={2835-8856},
year={2022},
url={https://openreview.net/forum?id=yzkSU5zdwD},
note={Survey Certification}
}

@article{JMLR:v25:23-0870,
  author  = {Hyung Won Chung and Le Hou and Shayne Longpre and Barret Zoph and Yi Tay and William Fedus and Yunxuan Li and Xuezhi Wang and Mostafa Dehghani and Siddhartha Brahma and Albert Webson and Shixiang Shane Gu and Zhuyun Dai and Mirac Suzgun and Xinyun Chen and Aakanksha Chowdhery and Alex Castro-Ros and Marie Pellat and Kevin Robinson and Dasha Valter and Sharan Narang and Gaurav Mishra and Adams Yu and Vincent Zhao and Yanping Huang and Andrew Dai and Hongkun Yu and Slav Petrov and Ed H. Chi and Jeff Dean and Jacob Devlin and Adam Roberts and Denny Zhou and Quoc V. Le and Jason Wei},
  title   = {{S}caling {I}nstruction-{F}inetuned {L}anguage Models},
  journal = {Journal of Machine Learning Research},
  year    = {2024},
  volume  = {25},
  number  = {70},
  pages   = {1--53},
  url     = {http://jmlr.org/papers/v25/23-0870.html}
}

@misc{thompson2022computationallimitsdeeplearning,
      title={{T}he {C}omputational {L}imits of {D}eep {L}earning}, 
      author={Neil C. Thompson and Kristjan Greenewald and Keeheon Lee and Gabriel F. Manso},
      year={2022},
      eprint={2007.05558},
      archivePrefix={arXiv},
      primaryClass={cs.LG},
      url={https://arxiv.org/abs/2007.05558}, 
}

@inproceedings{MLSYS2022_462211f6,
 author = {Wu, Carole-Jean and Raghavendra, Ramya and Gupta, Udit and Acun, Bilge and Ardalani, Newsha and Maeng, Kiwan and Chang, Gloria and Aga, Fiona and Huang, Jinshi and Bai, Charles and Gschwind, Michael and Gupta, Anurag and Ott, Myle and Melnikov, Anastasia and Candido, Salvatore and Brooks, David and Chauhan, Geeta and Lee, Benjamin and Lee, Hsien-Hsin and Akyildiz, Bugra and Balandat, Maximilian and Spisak, Joe and Jain, Ravi and Rabbat, Mike and Hazelwood, Kim},
 booktitle = {Proceedings of Machine Learning and Systems},
 editor = {D. Marculescu and Y. Chi and C. Wu},
 pages = {795--813},
 title = {{S}ustainable {AI}: {E}nvironmental {I}mplications, {C}hallenges and {O}pportunities},
 url = {https://proceedings.mlsys.org/paper_files/paper/2022/file/462211f67c7d858f663355eff93b745e-Paper.pdf},
 volume = {4},
 year = {2022}
}

@article{JMLR:v24:23-0069,
  author  = {Alexandra Sasha Luccioni and Sylvain Viguier and Anne-Laure Ligozat},
  title   = {{E}stimating the {C}arbon {F}ootprint of {BLOOM}, a 176{B} {P}arameter {L}anguage {M}odel},
  journal = {Journal of Machine Learning Research},
  year    = {2023},
  volume  = {24},
  number  = {253},
  pages   = {1--15},
  url     = {http://jmlr.org/papers/v24/23-0069.html}
}

@misc{villalobos2024rundatalimitsllm,
      title={{W}ill we run out of data? {L}imits of {LLM} scaling based on human-generated data}, 
      author={Pablo Villalobos and Anson Ho and Jaime Sevilla and Tamay Besiroglu and Lennart Heim and Marius Hobbhahn},
      year={2024},
      eprint={2211.04325},
      archivePrefix={arXiv},
      primaryClass={cs.LG},
      url={https://arxiv.org/abs/2211.04325}, 
}

@inproceedings{10.1145/3442188.3445922,
author = {Bender, Emily M. and Gebru, Timnit and McMillan-Major, Angelina and Shmitchell, Shmargaret},
title = {{O}n the {D}angers of {S}tochastic {P}arrots: {C}an {L}anguage {M}odels {B}e {T}oo {B}ig?},
year = {2021},
isbn = {9781450383097},
publisher = {Association for Computing Machinery},
address = {New York, NY, USA},
url = {https://doi.org/10.1145/3442188.3445922},
doi = {10.1145/3442188.3445922},
booktitle = {Proceedings of the 2021 ACM Conference on Fairness, Accountability, and Transparency},
pages = {610–623},
numpages = {14},
location = {Virtual Event, Canada},
series = {FAccT '21}
}

@inproceedings{NEURIPS2023_42f22550,
 author = {Birhane, Abeba and prabhu, vinay and Han, Sanghyun and Boddeti, Vishnu and Luccioni, Sasha},
 booktitle = {Advances in Neural Information Processing Systems},
 editor = {A. Oh and T. Naumann and A. Globerson and K. Saenko and M. Hardt and S. Levine},
 pages = {21268--21284},
 publisher = {Curran Associates, Inc.},
 title = {{I}nto the {LAION}’s {D}en: {I}nvestigating {H}ate in {M}ultimodal {D}atasets},
 url = {https://proceedings.neurips.cc/paper_files/paper/2023/file/42f225509e8263e2043c9d834ccd9a2b-Paper-Datasets_and_Benchmarks.pdf},
 volume = {36},
 year = {2023}
}

@inproceedings{kim-etal-2024-solar,
    title = "{SOLAR} 10.7{B}: {S}caling {L}arge {L}anguage {M}odels with {S}imple yet {E}ffective {D}epth {U}p-{S}caling",
    author = "Kim, Sanghoon  and
      Kim, Dahyun  and
      Park, Chanjun  and
      Lee, Wonsung  and
      Song, Wonho  and
      Kim, Yunsu  and
      Kim, Hyeonwoo  and
      Kim, Yungi  and
      Lee, Hyeonju  and
      Kim, Jihoo  and
      Ahn, Changbae  and
      Yang, Seonghoon  and
      Lee, Sukyung  and
      Park, Hyunbyung  and
      Gim, Gyoungjin  and
      Cha, Mikyoung  and
      Lee, Hwalsuk  and
      Kim, Sunghun",
    editor = "Yang, Yi  and
      Davani, Aida  and
      Sil, Avi  and
      Kumar, Anoop",
    booktitle = "Proceedings of the 2024 Conference of the North American Chapter of the Association for Computational Linguistics: Human Language Technologies (Volume 6: Industry Track)",
    month = jun,
    year = "2024",
    address = "Mexico City, Mexico",
    publisher = "Association for Computational Linguistics",
    url = "https://aclanthology.org/2024.naacl-industry.3/",
    doi = "10.18653/v1/2024.naacl-industry.3",
    pages = "23--35"
}

@inproceedings{NEURIPS2024_143ea4a1,
 author = {Du, Wenyu and Luo, Tongxu and Qiu, Zihan and Huang, Zeyu and Shen, Yikang and Cheng, Reynold and Guo, Yike and Fu, Jie},
 booktitle = {Advances in Neural Information Processing Systems},
 editor = {A. Globerson and L. Mackey and D. Belgrave and A. Fan and U. Paquet and J. Tomczak and C. Zhang},
 pages = {10491--10540},
 publisher = {Curran Associates, Inc.},
 title = {{S}tacking {Y}our {T}ransformers: {A} {C}loser {L}ook at {M}odel {G}rowth for {E}fficient {LLM} {P}re-{T}raining},
 url = {https://proceedings.neurips.cc/paper_files/paper/2024/file/143ea4a156ef64f32d4d905206cf32e1-Paper-Conference.pdf},
 volume = {37},
 year = {2024}
}

@InProceedings{pmlr-v162-shen22f,
  title = 	 {{S}taged {T}raining for {T}ransformer {L}anguage {M}odels},
  author =       {Shen, Sheng and Walsh, Pete and Keutzer, Kurt and Dodge, Jesse and Peters, Matthew and Beltagy, Iz},
  booktitle = 	 {Proceedings of the 39th International Conference on Machine Learning},
  pages = 	 {19893--19908},
  year = 	 {2022},
  editor = 	 {Chaudhuri, Kamalika and Jegelka, Stefanie and Song, Le and Szepesvari, Csaba and Niu, Gang and Sabato, Sivan},
  volume = 	 {162},
  series = 	 {Proceedings of Machine Learning Research},
  month = 	 {17--23 Jul},
  publisher =    {PMLR},
  url = 	 {https://proceedings.mlr.press/v162/shen22f.html}
}

@inproceedings{
wang2023learning,
title={{L}earning to {G}row {P}retrained {M}odels for {E}fficient {T}ransformer {T}raining},
author={Peihao Wang and Rameswar Panda and Lucas Torroba Hennigen and Philip Greengard and Leonid Karlinsky and Rogerio Feris and David Daniel Cox and Zhangyang Wang and Yoon Kim},
booktitle={International Conference on Learning Representations},
year={2023},
url={https://openreview.net/forum?id=cDYRS5iZ16f}
}

@inproceedings{
wang2024lemon,
title={{LEMON}: {L}ossless model expansion},
author={Yite Wang and Jiahao Su and Hanlin Lu and Cong Xie and Tianyi Liu and Jianbo Yuan and Haibin Lin and Ruoyu Sun and Hongxia Yang},
booktitle={The Twelfth International Conference on Learning Representations},
year={2024},
url={https://openreview.net/forum?id=3Vw7DQqq7U}
}

@misc{yang2020progressivelystacking20multistage,
      title={{P}rogressively {S}tacking 2.0: {A} {M}ulti-stage {L}ayerwise {T}raining {M}ethod for {BERT} {T}raining {S}peedup}, 
      author={Cheng Yang and Shengnan Wang and Chao Yang and Yuechuan Li and Ru He and Jingqiao Zhang},
      year={2020},
      eprint={2011.13635},
      archivePrefix={arXiv},
      primaryClass={cs.CL},
      url={https://arxiv.org/abs/2011.13635}, 
}

@inproceedings{chen-etal-2022-bert2bert,
    title = "bert2{BERT}: {T}owards {R}eusable {P}retrained {L}anguage {M}odels",
    author = "Chen, Cheng  and
      Yin, Yichun  and
      Shang, Lifeng  and
      Jiang, Xin  and
      Qin, Yujia  and
      Wang, Fengyu  and
      Wang, Zhi  and
      Chen, Xiao  and
      Liu, Zhiyuan  and
      Liu, Qun",
    editor = "Muresan, Smaranda  and
      Nakov, Preslav  and
      Villavicencio, Aline",
    booktitle = "Proceedings of the 60th Annual Meeting of the Association for Computational Linguistics (Volume 1: Long Papers)",
    month = may,
    year = "2022",
    address = "Dublin, Ireland",
    publisher = "Association for Computational Linguistics",
    url = "https://aclanthology.org/2022.acl-long.151/",
    doi = "10.18653/v1/2022.acl-long.151",
    pages = "2134--2148"
}

@inproceedings{
yao2024masked,
title={{M}asked {S}tructural {G}rowth for 2x {F}aster {L}anguage {M}odel {P}re-training},
author={Yiqun Yao and Zheng Zhang and Jing Li and Yequan Wang},
booktitle={The Twelfth International Conference on Learning Representations},
year={2024},
url={https://openreview.net/forum?id=rL7xsg1aRn}
}

@inproceedings{wu-etal-2024-llama,
    title = "{LL}a{MA} {P}ro: {P}rogressive {LL}a{MA} with {B}lock {E}xpansion",
    author = "Wu, Chengyue  and
      Gan, Yukang  and
      Ge, Yixiao  and
      Lu, Zeyu  and
      Wang, Jiahao  and
      Feng, Ye  and
      Shan, Ying  and
      Luo, Ping",
    editor = "Ku, Lun-Wei  and
      Martins, Andre  and
      Srikumar, Vivek",
    booktitle = "Proceedings of the 62nd Annual Meeting of the Association for Computational Linguistics (Volume 1: Long Papers)",
    month = aug,
    year = "2024",
    address = "Bangkok, Thailand",
    publisher = "Association for Computational Linguistics",
    url = "https://aclanthology.org/2024.acl-long.352/",
    doi = "10.18653/v1/2024.acl-long.352",
    pages = "6518--6537"
}

@misc{grattafiori2024llama3herdmodels,
      title={{T}he {L}lama 3 {H}erd of {M}odels}, 
      author={Aaron Grattafiori and Abhimanyu Dubey and Abhinav Jauhri and Abhinav Pandey and Abhishek Kadian and Ahmad Al-Dahle and Aiesha Letman and Akhil Mathur and Alan Schelten and Alex Vaughan and et al.},
      year={2024},
      eprint={2407.21783},
      archivePrefix={arXiv},
      primaryClass={cs.AI},
      url={https://arxiv.org/abs/2407.21783}, 
}

@inproceedings{jawahar-etal-2019-bert,
    title = "What {D}oes {BERT} {L}earn about the {S}tructure of {L}anguage?",
    author = "Jawahar, Ganesh  and
      Sagot, Beno{\^i}t  and
      Seddah, Djam{\'e}",
    editor = "Korhonen, Anna  and
      Traum, David  and
      M{\`a}rquez, Llu{\'i}s",
    booktitle = "Proceedings of the 57th Annual Meeting of the Association for Computational Linguistics",
    month = jul,
    year = "2019",
    address = "Florence, Italy",
    publisher = "Association for Computational Linguistics",
    url = "https://aclanthology.org/P19-1356/",
    doi = "10.18653/v1/P19-1356",
    pages = "3651--3657"
}

@inproceedings{
dao2024flashattention,
title={{F}lash{A}ttention-2: {F}aster {A}ttention with {B}etter {P}arallelism and {W}ork {P}artitioning},
author={Tri Dao},
booktitle={The Twelfth International Conference on Learning Representations},
year={2024},
url={https://openreview.net/forum?id=mZn2Xyh9Ec}
}

@misc{clark2018thinksolvedquestionanswering,
      title={{T}hink you have {S}olved {Q}uestion {A}nswering? {T}ry {ARC}, the {AI}2 {R}easoning {C}hallenge}, 
      author={Peter Clark and Isaac Cowhey and Oren Etzioni and Tushar Khot and Ashish Sabharwal and Carissa Schoenick and Oyvind Tafjord},
      year={2018},
      eprint={1803.05457},
      archivePrefix={arXiv},
      primaryClass={cs.AI},
      url={https://arxiv.org/abs/1803.05457}, 
}

@inproceedings{ijcai2020p501,
  title     = {{L}ogi{QA}: {A} {C}hallenge {D}ataset for {M}achine {R}eading {C}omprehension with {L}ogical {R}easoning},
  author    = {Liu, Jian and Cui, Leyang and Liu, Hanmeng and Huang, Dandan and Wang, Yile and Zhang, Yue},
  booktitle = {Proceedings of the Twenty-Ninth International Joint Conference on
               Artificial Intelligence, {IJCAI-20}},
  publisher = {International Joint Conferences on Artificial Intelligence Organization},
  editor    = {Christian Bessiere},
  pages     = {3622--3628},
  year      = {2020},
  month     = {7},
  note      = {Main track},
  doi       = {10.24963/ijcai.2020/501},
  url       = {https://doi.org/10.24963/ijcai.2020/501},
}

@inproceedings{Bisk2020,
    author = {Yonatan Bisk and Rowan Zellers and
            Ronan Le Bras and Jianfeng Gao
            and Yejin Choi},
    title = {{PIQA}: {R}easoning about {P}hysical {C}ommonsense in
           {N}atural {L}anguage},
    booktitle = {Thirty-Fourth AAAI Conference on
               Artificial Intelligence},
    year = {2020},
    url={https://ojs.aaai.org/index.php/AAAI/article/view/6239}, 
}

@article{10.1145/3474381,
author = {Sakaguchi, Keisuke and Bras, Ronan Le and Bhagavatula, Chandra and Choi, Yejin},
title = {{W}ino{G}rande: an adversarial winograd schema challenge at scale},
year = {2021},
issue_date = {September 2021},
publisher = {Association for Computing Machinery},
address = {New York, NY, USA},
volume = {64},
number = {9},
issn = {0001-0782},
url = {https://doi.org/10.1145/3474381},
doi = {10.1145/3474381},
journal = {Commun. ACM},
month = aug,
pages = {99–106},
numpages = {8}
}

@inproceedings{
merity2017pointer,
title={{P}ointer {S}entinel {M}ixture {M}odels},
author={Stephen Merity and Caiming Xiong and James Bradbury and Richard Socher},
booktitle={International Conference on Learning Representations},
year={2017},
url={https://openreview.net/forum?id=Byj72udxe}
}

@misc{eval-harness,
  author       = {Gao, Leo and Tow, Jonathan and Abbasi, Baber and Biderman, Stella and Black, Sid and DiPofi, Anthony and Foster, Charles and Golding, Laurence and Hsu, Jeffrey and Le Noac'h, Alain and Li, Haonan and McDonell, Kyle and Muennighoff, Niklas and Ociepa, Chris and Phang, Jason and Reynolds, Laria and Schoelkopf, Hailey and Skowron, Aviya and Sutawika, Lintang and Tang, Eric and Thite, Anish and Wang, Ben and Wang, Kevin and Zou, Andy},
  title        = {A framework for few-shot language model evaluation},
  month        = 07,
  year         = 2024,
  publisher    = {Zenodo},
  version      = {v0.4.3},
  doi          = {10.5281/zenodo.12608602},
  url          = {https://zenodo.org/records/12608602}
}

@article{10.1137/060659624,
author = {Knight, Philip A.},
title = {{T}he {S}inkhorn-{K}nopp {A}lgorithm: {C}onvergence and {A}pplications},
year = {2008},
issue_date = {February 2008},
publisher = {Society for Industrial and Applied Mathematics},
address = {USA},
volume = {30},
number = {1},
issn = {0895-4798},
url = {https://doi.org/10.1137/060659624},
doi = {10.1137/060659624},
journal = {SIAM Journal on Matrix Analysis and Applications},
month = mar,
pages = {261–275},
numpages = {15}
}

@inproceedings{NEURIPS2020_fb269786,
 author = {Singh, Sidak Pal and Jaggi, Martin},
 booktitle = {Advances in Neural Information Processing Systems},
 editor = {H. Larochelle and M. Ranzato and R. Hadsell and M.F. Balcan and H. Lin},
 pages = {22045--22055},
 publisher = {Curran Associates, Inc.},
 title = {{M}odel {F}usion via {O}ptimal {T}ransport},
 url = {https://proceedings.neurips.cc/paper_files/paper/2020/file/fb2697869f56484404c8ceee2985b01d-Paper.pdf},
 volume = {33},
 year = {2020}
}

@inproceedings{
imfeld2024transformer,
title={{T}ransformer {F}usion with {O}ptimal {T}ransport},
author={Moritz Imfeld and Jacopo Graldi and Marco Giordano and Thomas Hofmann and Sotiris Anagnostidis and Sidak Pal Singh},
booktitle={The Twelfth International Conference on Learning Representations},
year={2024},
url={https://openreview.net/forum?id=LjeqMvQpen}
}

@inproceedings{yang-etal-2025-lesa,
    title = "{LESA}: Learnable {LLM} Layer Scaling-Up",
    author = "Yang, Yifei  and
      Cao, Zouying  and
      Ma, Xinbei  and
      Yao, Yao  and
      Chen, Zhi  and
      Qin, Libo  and
      Zhao, Hai",
    editor = "Che, Wanxiang  and
      Nabende, Joyce  and
      Shutova, Ekaterina  and
      Pilehvar, Mohammad Taher",
    booktitle = "Proceedings of the 63rd Annual Meeting of the Association for Computational Linguistics (Volume 1: Long Papers)",
    month = jul,
    year = "2025",
    address = "Vienna, Austria",
    publisher = "Association for Computational Linguistics",
    url = "https://aclanthology.org/2025.acl-long.1095/",
    pages = "22463--22476",
    ISBN = "979-8-89176-251-0"
}

@inproceedings{NEURIPS2024_370df50c,
 author = {Penedo, Guilherme and Kydl\'{\i}\v{c}ek, Hynek and allal, Loubna Ben and Lozhkov, Anton and Mitchell, Margaret and Raffel, Colin and Von Werra, Leandro and Wolf, Thomas},
 booktitle = {Advances in Neural Information Processing Systems},
 editor = {A. Globerson and L. Mackey and D. Belgrave and A. Fan and U. Paquet and J. Tomczak and C. Zhang},
 pages = {30811--30849},
 publisher = {Curran Associates, Inc.},
 title = {{T}he {F}ineWeb {D}atasets: {D}ecanting the {W}eb for the {F}inest {T}ext {D}ata at {S}cale},
 url = {https://proceedings.neurips.cc/paper_files/paper/2024/file/370df50ccfdf8bde18f8f9c2d9151bda-Paper-Datasets_and_Benchmarks_Track.pdf},
 volume = {37},
 year = {2024}
}

@InProceedings{pmlr-v97-gong19a,
  title = 	 {{E}fficient {T}raining of {BERT} by {P}rogressively {S}tacking},
  author =       {Gong, Linyuan and He, Di and Li, Zhuohan and Qin, Tao and Wang, Liwei and Liu, Tieyan},
  booktitle = 	 {Proceedings of the 36th International Conference on Machine Learning},
  pages = 	 {2337--2346},
  year = 	 {2019},
  editor = 	 {Chaudhuri, Kamalika and Salakhutdinov, Ruslan},
  volume = 	 {97},
  series = 	 {Proceedings of Machine Learning Research},
  month = 	 {09--15 Jun},
  publisher =    {PMLR},
  url = 	 {https://proceedings.mlr.press/v97/gong19a.html}
}

@inproceedings{Chen2016Net2Net,
    author = {Chen, Tianqi and Goodfellow, Ian and Shlens, Jonathon},
    title = {{N}et2{N}et: {A}ccelerating {L}earning via {K}nowledge {T}ransfer},
    booktitle = {International Conference on Learning Representations},
    year = {2016},
    url = {https://arxiv.org/abs/1511.05641}
}

@inproceedings{pmlr-v48-wei16,
    author = {Wei, Tao and Wang, Changhu and Rui, Yong and Chen, Chang Wen},
    title = {{N}etwork {M}orphism},
    booktitle = {Proceedings of The 33rd International Conference on Machine Learning},
    volume = {48},
    pages = {564--572},
    year = {2016},
    publisher = {PMLR},
    series = {Proceedings of Machine Learning Research},
    editor = {Balcan, Maria Florina and Weinberger, Kilian Q.},
    url = {https://proceedings.mlr.press/v48/wei16.html}
}

@misc{rusu2022progressiveneuralnetworks,
      title={{P}rogressive {N}eural {N}etworks}, 
      author={Andrei A. Rusu and Neil C. Rabinowitz and Guillaume Desjardins and Hubert Soyer and James Kirkpatrick and Koray Kavukcuoglu and Razvan Pascanu and Raia Hadsell},
      year={2022},
      eprint={1606.04671},
      archivePrefix={arXiv},
      primaryClass={cs.LG},
      url={https://arxiv.org/abs/1606.04671}, 
}

@inproceedings{
chang2018multilevel,
title={{M}ulti-level {R}esidual {N}etworks from {D}ynamical {S}ystems {V}iew},
author={Bo Chang and Lili Meng and Eldad Haber and Frederick Tung and David Begert},
booktitle={International Conference on Learning Representations},
year={2018},
url={https://openreview.net/forum?id=SyJS-OgR-},
}

@inproceedings{clark-etal-2019-boolq,
    title = "{B}ool{Q}: {E}xploring the {S}urprising {D}ifficulty of {N}atural {Y}es/{N}o {Q}uestions",
    author = "Clark, Christopher  and
      Lee, Kenton  and
      Chang, Ming-Wei  and
      Kwiatkowski, Tom  and
      Collins, Michael  and
      Toutanova, Kristina",
    editor = "Burstein, Jill  and
      Doran, Christy  and
      Solorio, Thamar",
    booktitle = "Proceedings of the 2019 Conference of the North {A}merican Chapter of the Association for Computational Linguistics: Human Language Technologies, Volume 1 (Long and Short Papers)",
    month = jun,
    year = "2019",
    address = "Minneapolis, Minnesota",
    publisher = "Association for Computational Linguistics",
    url = "https://aclanthology.org/N19-1300/",
    doi = "10.18653/v1/N19-1300",
    pages = "2924--2936"
}

@inproceedings{talmor-etal-2019-commonsenseqa,
    title = "{C}ommonsense{QA}: {A} {Q}uestion {A}nswering {C}hallenge {T}argeting {C}ommonsense {K}nowledge",
    author = "Talmor, Alon  and
      Herzig, Jonathan  and
      Lourie, Nicholas  and
      Berant, Jonathan",
    editor = "Burstein, Jill  and
      Doran, Christy  and
      Solorio, Thamar",
    booktitle = "Proceedings of the 2019 Conference of the North {A}merican Chapter of the Association for Computational Linguistics: Human Language Technologies, Volume 1 (Long and Short Papers)",
    month = jun,
    year = "2019",
    address = "Minneapolis, Minnesota",
    publisher = "Association for Computational Linguistics",
    url = "https://aclanthology.org/N19-1421/",
    doi = "10.18653/v1/N19-1421",
    pages = "4149--4158"
}

@inproceedings{
hendrycks2021measuring,
title={{M}easuring {M}assive {M}ultitask {L}anguage {U}nderstanding},
author={Dan Hendrycks and Collin Burns and Steven Basart and Andy Zou and Mantas Mazeika and Dawn Song and Jacob Steinhardt},
booktitle={International Conference on Learning Representations},
year={2021},
url={https://openreview.net/forum?id=d7KBjmI3GmQ}
}

@inproceedings{yano-etal-2025-step,
    title = "{STEP}: {S}taged {P}arameter-{E}fficient {P}re-training for {L}arge {L}anguage {M}odels",
    author = "Yano, Kazuki  and
      Ito, Takumi  and
      Suzuki, Jun",
    editor = "Chiruzzo, Luis  and
      Ritter, Alan  and
      Wang, Lu",
    booktitle = "Proceedings of the 2025 Conference of the Nations of the Americas Chapter of the Association for Computational Linguistics: Human Language Technologies (Volume 2: Short Papers)",
    month = apr,
    year = "2025",
    address = "Albuquerque, New Mexico",
    publisher = "Association for Computational Linguistics",
    url = "https://aclanthology.org/2025.naacl-short.32/",
    pages = "374--384",
    ISBN = "979-8-89176-190-2"
}

@InProceedings{pmlr-v44-li15convergent,
  title = 	 {{C}onvergent {L}earning: {D}o different neural networks learn the same representations?},
  author = 	 {Li, Yixuan and Yosinski, Jason and Clune, Jeff and Lipson, Hod and Hopcroft, John},
  booktitle = 	 {Proceedings of the 1st International Workshop on Feature Extraction: Modern Questions and Challenges at NIPS 2015},
  pages = 	 {196--212},
  year = 	 {2015},
  editor = 	 {Storcheus, Dmitry and Rostamizadeh, Afshin and Kumar, Sanjiv},
  volume = 	 {44},
  series = 	 {Proceedings of Machine Learning Research},
  address = 	 {Montreal, Canada},
  month = 	 {11 Dec},
  publisher =    {PMLR},
  url = 	 {https://proceedings.mlr.press/v44/li15convergent.html}
}

@InProceedings{pmlr-v97-yurochkin19a,
  title = 	 {{B}ayesian {N}onparametric {F}ederated {L}earning of {N}eural {N}etworks},
  author =       {Yurochkin, Mikhail and Agarwal, Mayank and Ghosh, Soumya and Greenewald, Kristjan and Hoang, Nghia and Khazaeni, Yasaman},
  booktitle = 	 {Proceedings of the 36th International Conference on Machine Learning},
  pages = 	 {7252--7261},
  year = 	 {2019},
  editor = 	 {Chaudhuri, Kamalika and Salakhutdinov, Ruslan},
  volume = 	 {97},
  series = 	 {Proceedings of Machine Learning Research},
  month = 	 {09--15 Jun},
  publisher =    {PMLR},
  url = 	 {https://proceedings.mlr.press/v97/yurochkin19a.html}
}

@inproceedings{NEURIPS2019_ecb287ff,
 author = {Yurochkin, Mikhail and Agarwal, Mayank and Ghosh, Soumya and Greenewald, Kristjan and Hoang, Nghia},
 booktitle = {Advances in Neural Information Processing Systems},
 editor = {H. Wallach and H. Larochelle and A. Beygelzimer and F. d\textquotesingle Alch\'{e}-Buc and E. Fox and R. Garnett},
 pages = {},
 publisher = {Curran Associates, Inc.},
 title = {{S}tatistical {M}odel {A}ggregation via {P}arameter {M}atching},
 url = {https://proceedings.neurips.cc/paper_files/paper/2019/file/ecb287ff763c169694f682af52c1f309-Paper.pdf},
 volume = {32},
 year = {2019}
}

@misc{peng2023instructiontuninggpt4,
      title={Instruction Tuning with GPT-4}, 
      author={Baolin Peng and Chunyuan Li and Pengcheng He and Michel Galley and Jianfeng Gao},
      year={2023},
      eprint={2304.03277},
      archivePrefix={arXiv},
      primaryClass={cs.CL},
      url={https://arxiv.org/abs/2304.03277}, 
}

@inproceedings{men-etal-2025-shortgpt,
    title = "{S}hort{GPT}: Layers in Large Language Models are More Redundant Than You Expect",
    author = "Men, Xin  and
      Xu, Mingyu  and
      Zhang, Qingyu  and
      Yuan, Qianhao  and
      Wang, Bingning  and
      Lin, Hongyu  and
      Lu, Yaojie  and
      Han, Xianpei  and
      Chen, Weipeng",
    editor = "Che, Wanxiang  and
      Nabende, Joyce  and
      Shutova, Ekaterina  and
      Pilehvar, Mohammad Taher",
    booktitle = "Findings of the Association for Computational Linguistics: ACL 2025",
    month = jul,
    year = "2025",
    address = "Vienna, Austria",
    publisher = "Association for Computational Linguistics",
    url = "https://aclanthology.org/2025.findings-acl.1035/",
    pages = "20192--20204",
    ISBN = "979-8-89176-256-5"
}

@misc{nguyen2016multifacetedfeaturevisualizationuncovering,
      title={Multifaceted Feature Visualization: Uncovering the Different Types of Features Learned By Each Neuron in Deep Neural Networks}, 
      author={Anh Nguyen and Jason Yosinski and Jeff Clune},
      year={2016},
      eprint={1602.03616},
      archivePrefix={arXiv},
      primaryClass={cs.NE},
      url={https://arxiv.org/abs/1602.03616}, 
}

@article{10.1145/3728458,
author = {Klabunde, Max and Schumacher, Tobias and Strohmaier, Markus and Lemmerich, Florian},
title = {Similarity of Neural Network Models: A Survey of Functional and Representational Measures},
year = {2025},
issue_date = {September 2025},
publisher = {Association for Computing Machinery},
address = {New York, NY, USA},
volume = {57},
number = {9},
issn = {0360-0300},
url = {https://doi.org/10.1145/3728458},
doi = {10.1145/3728458},
journal = {ACM Comput. Surv.},
month = may,
articleno = {242},
numpages = {52}
}

@article{sajjad-etal-2022-neuron,
    title = "Neuron-level Interpretation of Deep {NLP} Models: A Survey",
    author = "Sajjad, Hassan  and
      Durrani, Nadir  and
      Dalvi, Fahim",
    editor = "Roark, Brian  and
      Nenkova, Ani",
    journal = "Transactions of the Association for Computational Linguistics",
    volume = "10",
    year = "2022",
    address = "Cambridge, MA",
    publisher = "MIT Press",
    url = "https://aclanthology.org/2022.tacl-1.74/",
    doi = "10.1162/tacl_a_00519",
    pages = "1285--1303"
}

@inproceedings{NIPS2013_af21d0c9,
 author = {Cuturi, Marco},
 booktitle = {Advances in Neural Information Processing Systems},
 editor = {C.J. Burges and L. Bottou and M. Welling and Z. Ghahramani and K.Q. Weinberger},
 pages = {},
 publisher = {Curran Associates, Inc.},
 title = {Sinkhorn Distances: Lightspeed Computation of Optimal Transport},
 url = {https://proceedings.neurips.cc/paper_files/paper/2013/file/af21d0c97db2e27e13572cbf59eb343d-Paper.pdf},
 volume = {26},
 year = {2013}
}

@Article{e19090474,
AUTHOR = {Tax, Tycho M.S. and Mediano, Pedro A.M. and Shanahan, Murray},
TITLE = {The Partial Information Decomposition of Generative Neural Network Models},
JOURNAL = {Entropy},
VOLUME = {19},
YEAR = {2017},
NUMBER = {9},
ARTICLE-NUMBER = {474},
URL = {https://www.mdpi.com/1099-4300/19/9/474},
ISSN = {1099-4300},
DOI = {10.3390/e19090474}
}

@ARTICLE{8998186,
  author={Yu, Shujian and Wickstrøm, Kristoffer and Jenssen, Robert and Príncipe, José C.},
  journal={IEEE Transactions on Neural Networks and Learning Systems}, 
  title={Understanding Convolutional Neural Networks With Information Theory: An Initial Exploration}, 
  year={2021},
  volume={32},
  number={1},
  pages={435-442},
  doi={10.1109/TNNLS.2020.2968509}}

@inproceedings{williams2025compressinglanguagemodelsspecialized,
    title = "Compressing Language Models for Specialized Domains",
    author = "Williams, Miles  and
      Chrysostomou, George  and
      Jeronymo, Vitor Amancio  and
      Aletras, Nikolaos",
    editor = "Demberg, Vera  and
      Inui, Kentaro  and
      Marquez, Llu{\'i}s",
    booktitle = "Proceedings of the 19th Conference of the {E}uropean Chapter of the {A}ssociation for {C}omputational {L}inguistics (Volume 1: Long Papers)",
    month = mar,
    year = "2026",
    address = "Rabat, Morocco",
    publisher = "Association for Computational Linguistics",
    url = "https://aclanthology.org/2026.eacl-long.347/",
    doi = "10.18653/v1/2026.eacl-long.347",
    pages = "7393--7415",
    ISBN = "979-8-89176-380-7"
}

@misc{singhal2022largelanguagemodelsencode,
      title={Large Language Models Encode Clinical Knowledge}, 
      author={Karan Singhal and Shekoofeh Azizi and Tao Tu and S. Sara Mahdavi and Jason Wei and Hyung Won Chung and Nathan Scales and Ajay Tanwani and Heather Cole-Lewis and Stephen Pfohl and Perry Payne and Martin Seneviratne and Paul Gamble and Chris Kelly and Nathaneal Scharli and Aakanksha Chowdhery and Philip Mansfield and Blaise Aguera y Arcas and Dale Webster and Greg S. Corrado and Yossi Matias and Katherine Chou and Juraj Gottweis and Nenad Tomasev and Yun Liu and Alvin Rajkomar and Joelle Barral and Christopher Semturs and Alan Karthikesalingam and Vivek Natarajan},
      year={2022},
      eprint={2212.13138},
      archivePrefix={arXiv},
      primaryClass={cs.CL},
      url={https://arxiv.org/abs/2212.13138}, 
}

@inproceedings{jin-etal-2019-pubmedqa,
    title = "{P}ub{M}ed{QA}: A Dataset for Biomedical Research Question Answering",
    author = "Jin, Qiao  and
      Dhingra, Bhuwan  and
      Liu, Zhengping  and
      Cohen, William  and
      Lu, Xinghua",
    editor = "Inui, Kentaro  and
      Jiang, Jing  and
      Ng, Vincent  and
      Wan, Xiaojun",
    booktitle = "Proceedings of the 2019 Conference on Empirical Methods in Natural Language Processing and the 9th International Joint Conference on Natural Language Processing (EMNLP-IJCNLP)",
    month = nov,
    year = "2019",
    address = "Hong Kong, China",
    publisher = "Association for Computational Linguistics",
    url = "https://aclanthology.org/D19-1259/",
    doi = "10.18653/v1/D19-1259",
    pages = "2567--2577"
}

@Article{app11146421,
AUTHOR = {Jin, Di and Pan, Eileen and Oufattole, Nassim and Weng, Wei-Hung and Fang, Hanyi and Szolovits, Peter},
TITLE = {What Disease Does This Patient Have? A Large-Scale Open Domain Question Answering Dataset from Medical Exams},
JOURNAL = {Applied Sciences},
VOLUME = {11},
YEAR = {2021},
NUMBER = {14},
ARTICLE-NUMBER = {6421},
URL = {https://www.mdpi.com/2076-3417/11/14/6421},
ISSN = {2076-3417},
DOI = {10.3390/app11146421}
}

@InProceedings{pmlr-v174-pal22a,
  title = 	 {MedMCQA: A Large-scale Multi-Subject Multi-Choice Dataset for Medical domain Question Answering},
  author =       {Pal, Ankit and Umapathi, Logesh Kumar and Sankarasubbu, Malaikannan},
  booktitle = 	 {Proceedings of the Conference on Health, Inference, and Learning},
  pages = 	 {248--260},
  year = 	 {2022},
  editor = 	 {Flores, Gerardo and Chen, George H and Pollard, Tom and Ho, Joyce C and Naumann, Tristan},
  volume = 	 {174},
  series = 	 {Proceedings of Machine Learning Research},
  month = 	 {07--08 Apr},
  publisher =    {PMLR},
  url = 	 {https://proceedings.mlr.press/v174/pal22a.html}
}

@inproceedings{10.1145/3462757.3466088,
author = {Zheng, Lucia and Guha, Neel and Anderson, Brandon R. and Henderson, Peter and Ho, Daniel E.},
title = {When does pretraining help? assessing self-supervised learning for law and the CaseHOLD dataset of 53,000+ legal holdings},
year = {2021},
isbn = {9781450385268},
publisher = {Association for Computing Machinery},
address = {New York, NY, USA},
url = {https://doi.org/10.1145/3462757.3466088},
doi = {10.1145/3462757.3466088},
booktitle = {Proceedings of the Eighteenth International Conference on Artificial Intelligence and Law},
pages = {159–168},
numpages = {10},
location = {S\~{a}o Paulo, Brazil},
series = {ICAIL '21}
}

@inproceedings{chalkidis-etal-2022-lexglue,
    title = "{L}ex{GLUE}: A Benchmark Dataset for Legal Language Understanding in {E}nglish",
    author = "Chalkidis, Ilias  and
      Jana, Abhik  and
      Hartung, Dirk  and
      Bommarito, Michael  and
      Androutsopoulos, Ion  and
      Katz, Daniel  and
      Aletras, Nikolaos",
    editor = "Muresan, Smaranda  and
      Nakov, Preslav  and
      Villavicencio, Aline",
    booktitle = "Proceedings of the 60th Annual Meeting of the Association for Computational Linguistics (Volume 1: Long Papers)",
    month = may,
    year = "2022",
    address = "Dublin, Ireland",
    publisher = "Association for Computational Linguistics",
    url = "https://aclanthology.org/2022.acl-long.297/",
    doi = "10.18653/v1/2022.acl-long.297",
    pages = "4310--4330"
}

@inproceedings{chalkidis-etal-2019-neural,
    title = "Neural Legal Judgment Prediction in {E}nglish",
    author = "Chalkidis, Ilias  and
      Androutsopoulos, Ion  and
      Aletras, Nikolaos",
    editor = "Korhonen, Anna  and
      Traum, David  and
      M{\`a}rquez, Llu{\'i}s",
    booktitle = "Proceedings of the 57th Annual Meeting of the Association for Computational Linguistics",
    month = jul,
    year = "2019",
    address = "Florence, Italy",
    publisher = "Association for Computational Linguistics",
    url = "https://aclanthology.org/P19-1424/",
    doi = "10.18653/v1/P19-1424",
    pages = "4317--4323"
}

@article{10.1093/bioinformatics/btz682,
    author = {Lee, Jinhyuk and Yoon, Wonjin and Kim, Sungdong and Kim, Donghyeon and Kim, Sunkyu and So, Chan Ho and Kang, Jaewoo},
    title = {BioBERT: a pre-trained biomedical language representation model for biomedical text mining},
    journal = {Bioinformatics},
    volume = {36},
    number = {4},
    pages = {1234-1240},
    year = {2019},
    month = {09},
    issn = {1367-4803},
    doi = {10.1093/bioinformatics/btz682},
    url = {https://doi.org/10.1093/bioinformatics/btz682},
    eprint = {https://academic.oup.com/bioinformatics/article-pdf/36/4/1234/48983216/bioinformatics_36_4_1234.pdf},
}

@article{10.1145/3458754,
author = {Gu, Yu and Tinn, Robert and Cheng, Hao and Lucas, Michael and Usuyama, Naoto and Liu, Xiaodong and Naumann, Tristan and Gao, Jianfeng and Poon, Hoifung},
title = {Domain-Specific Language Model Pretraining for Biomedical Natural Language Processing},
year = {2021},
issue_date = {January 2022},
publisher = {Association for Computing Machinery},
address = {New York, NY, USA},
volume = {3},
number = {1},
url = {https://doi.org/10.1145/3458754},
doi = {10.1145/3458754},
journal = {ACM Trans. Comput. Healthcare},
month = oct,
articleno = {2},
numpages = {23}
}

@article{10.1093/bib/bbac409,
    author = {Luo, Renqian and Sun, Liai and Xia, Yingce and Qin, Tao and Zhang, Sheng and Poon, Hoifung and Liu, Tie-Yan},
    title = {BioGPT: generative pre-trained transformer for biomedical text generation and mining},
    journal = {Briefings in Bioinformatics},
    volume = {23},
    number = {6},
    pages = {bbac409},
    year = {2022},
    month = {09},
    issn = {1477-4054},
    doi = {10.1093/bib/bbac409},
    url = {https://doi.org/10.1093/bib/bbac409},
    eprint = {https://academic.oup.com/bib/article-pdf/23/6/bbac409/47144271/bbac409.pdf},
}

@misc{singhal2023expertlevelmedicalquestionanswering,
      title={Towards Expert-Level Medical Question Answering with Large Language Models}, 
      author={Karan Singhal and Tao Tu and Juraj Gottweis and Rory Sayres and Ellery Wulczyn and Le Hou and Kevin Clark and Stephen Pfohl and Heather Cole-Lewis and Darlene Neal and Mike Schaekermann and Amy Wang and Mohamed Amin and Sami Lachgar and Philip Mansfield and Sushant Prakash and Bradley Green and Ewa Dominowska and Blaise Aguera y Arcas and Nenad Tomasev and Yun Liu and Renee Wong and Christopher Semturs and S. Sara Mahdavi and Joelle Barral and Dale Webster and Greg S. Corrado and Yossi Matias and Shekoofeh Azizi and Alan Karthikesalingam and Vivek Natarajan},
      year={2023},
      eprint={2305.09617},
      archivePrefix={arXiv},
      primaryClass={cs.CL},
      url={https://arxiv.org/abs/2305.09617}, 
}

@inproceedings{NEURIPS2022_bc218a0c,
 author = {Henderson, Peter and Krass, Mark and Zheng, Lucia and Guha, Neel and Manning, Christopher D and Jurafsky, Dan and Ho, Daniel},
 booktitle = {Advances in Neural Information Processing Systems},
 editor = {S. Koyejo and S. Mohamed and A. Agarwal and D. Belgrave and K. Cho and A. Oh},
 pages = {29217--29234},
 publisher = {Curran Associates, Inc.},
 title = {Pile of Law: Learning Responsible Data Filtering from the Law and a 256GB Open-Source Legal Dataset},
 url = {https://proceedings.neurips.cc/paper_files/paper/2022/file/bc218a0c656e49d4b086975a9c785f47-Paper-Datasets_and_Benchmarks.pdf},
 volume = {35},
 year = {2022}
}

@inproceedings{niklaus-etal-2024-multilegalpile,
    title = "{M}ulti{L}egal{P}ile: A 689{GB} Multilingual Legal Corpus",
    author = {Niklaus, Joel  and
      Matoshi, Veton  and
      St{\"u}rmer, Matthias  and
      Chalkidis, Ilias  and
      Ho, Daniel},
    editor = "Ku, Lun-Wei  and
      Martins, Andre  and
      Srikumar, Vivek",
    booktitle = "Proceedings of the 62nd Annual Meeting of the Association for Computational Linguistics (Volume 1: Long Papers)",
    month = aug,
    year = "2024",
    address = "Bangkok, Thailand",
    publisher = "Association for Computational Linguistics",
    url = "https://aclanthology.org/2024.acl-long.805/",
    doi = "10.18653/v1/2024.acl-long.805",
    pages = "15077--15094"
}

@inproceedings{pires-etal-2023-one,
    title = "One Wide Feedforward Is All You Need",
    author = "Pires, Telmo  and
      Vilarinho Lopes, Ant{\'o}nio  and
      Assogba, Yannick  and
      Setiawan, Hendra",
    editor = "Koehn, Philipp  and
      Haddow, Barry  and
      Kocmi, Tom  and
      Monz, Christof",
    booktitle = "Proceedings of the Eighth Conference on Machine Translation",
    month = dec,
    year = "2023",
    address = "Singapore",
    publisher = "Association for Computational Linguistics",
    url = "https://aclanthology.org/2023.wmt-1.98/",
    doi = "10.18653/v1/2023.wmt-1.98",
    pages = "1031--1044"
}

@misc{wei2024investigatinglowranktrainingtransformer,
      title={Investigating Low-Rank Training in Transformer Language Models: Efficiency and Scaling Analysis}, 
      author={Xiuying Wei and Skander Moalla and Razvan Pascanu and Caglar Gulcehre},
      year={2024},
      eprint={2407.09835},
      archivePrefix={arXiv},
      primaryClass={cs.CL},
      url={https://arxiv.org/abs/2407.09835}, 
}

@InProceedings{pmlr-v119-bhojanapalli20a,
  title = 	 {Low-Rank Bottleneck in Multi-head Attention Models},
  author =       {Bhojanapalli, Srinadh and Yun, Chulhee and Rawat, Ankit Singh and Reddi, Sashank and Kumar, Sanjiv},
  booktitle = 	 {Proceedings of the 37th International Conference on Machine Learning},
  pages = 	 {864--873},
  year = 	 {2020},
  editor = 	 {III, Hal Daumé and Singh, Aarti},
  volume = 	 {119},
  series = 	 {Proceedings of Machine Learning Research},
  month = 	 {13--18 Jul},
  publisher =    {PMLR},
  url = 	 {https://proceedings.mlr.press/v119/bhojanapalli20a.html}
}

@misc{shazeer2020gluvariantsimprovetransformer,
      title={GLU Variants Improve Transformer}, 
      author={Noam Shazeer},
      year={2020},
      eprint={2002.05202},
      archivePrefix={arXiv},
      primaryClass={cs.LG},
      url={https://arxiv.org/abs/2002.05202}, 
}

@inproceedings{ainslie-etal-2023-gqa,
    title = "{GQA}: Training Generalized Multi-Query Transformer Models from Multi-Head Checkpoints",
    author = "Ainslie, Joshua  and
      Lee-Thorp, James  and
      de Jong, Michiel  and
      Zemlyanskiy, Yury  and
      Lebron, Federico  and
      Sanghai, Sumit",
    editor = "Bouamor, Houda  and
      Pino, Juan  and
      Bali, Kalika",
    booktitle = "Proceedings of the 2023 Conference on Empirical Methods in Natural Language Processing",
    month = dec,
    year = "2023",
    address = "Singapore",
    publisher = "Association for Computational Linguistics",
    url = "https://aclanthology.org/2023.emnlp-main.298/",
    doi = "10.18653/v1/2023.emnlp-main.298",
    pages = "4895--4901"
}

@ARTICLE{6773024,
  author={Shannon, C. E.},
  journal={The Bell System Technical Journal}, 
  title={A mathematical theory of communication}, 
  year={1948},
  volume={27},
  number={3},
  pages={379-423},
  doi={10.1002/j.1538-7305.1948.tb01338.x}}

@misc{yang2025qwen3technicalreport,
      title={Qwen3 Technical Report}, 
      author={An Yang and Anfeng Li and Baosong Yang and Beichen Zhang and Binyuan Hui and Bo Zheng and Bowen Yu and Chang Gao and Chengen Huang and Chenxu Lv and Chujie Zheng and Dayiheng Liu and Fan Zhou and Fei Huang and Feng Hu and Hao Ge and Haoran Wei and Huan Lin and Jialong Tang and Jian Yang and Jianhong Tu and Jianwei Zhang and Jianxin Yang and Jiaxi Yang and Jing Zhou and Jingren Zhou and Junyang Lin and Kai Dang and Keqin Bao and Kexin Yang and Le Yu and Lianghao Deng and Mei Li and Mingfeng Xue and Mingze Li and Pei Zhang and Peng Wang and Qin Zhu and Rui Men and Ruize Gao and Shixuan Liu and Shuang Luo and Tianhao Li and Tianyi Tang and Wenbiao Yin and Xingzhang Ren and Xinyu Wang and Xinyu Zhang and Xuancheng Ren and Yang Fan and Yang Su and Yichang Zhang and Yinger Zhang and Yu Wan and Yuqiong Liu and Zekun Wang and Zeyu Cui and Zhenru Zhang and Zhipeng Zhou and Zihan Qiu},
      year={2025},
      eprint={2505.09388},
      archivePrefix={arXiv},
      primaryClass={cs.CL},
      url={https://arxiv.org/abs/2505.09388}, 
}

@inproceedings{arabelly-etal-2025-improving,
    title = "Improving Task Diversity in Label Efficient Supervised Finetuning of {LLM}s",
    author = "Arabelly, Abhinav  and
      Nemade, Jagrut  and
      Nowak, Robert D  and
      Zhang, Jifan",
    editor = "Christodoulopoulos, Christos  and
      Chakraborty, Tanmoy  and
      Rose, Carolyn  and
      Peng, Violet",
    booktitle = "Proceedings of the 2025 Conference on Empirical Methods in Natural Language Processing",
    month = nov,
    year = "2025",
    address = "Suzhou, China",
    publisher = "Association for Computational Linguistics",
    url = "https://aclanthology.org/2025.emnlp-main.1093/",
    doi = "10.18653/v1/2025.emnlp-main.1093",
    pages = "21558--21570",
    ISBN = "979-8-89176-332-6"
}

@inproceedings{cao-etal-2026-moa,
    title = "{M}o{A}: Heterogeneous Mixture of Adapters for Parameter-Efficient Fine-Tuning of Large Language Models",
    author = "Cao, Jie  and
      Lin, Tianwei  and
      Yuan, Bo  and
      Yan, Rolan  and
      He, Hongyang  and
      Zhang, Wenqiao  and
      Li, Juncheng  and
      Zhang, Dongping  and
      Tang, Siliang  and
      Zhuang, Yueting",
    editor = "Liakata, Maria  and
      Moreira, Viviane P.  and
      Zhang, Jiajun  and
      Jurgens, David",
    booktitle = "Proceedings of the 64th Annual Meeting of the {A}ssociation for {C}omputational {L}inguistics (Volume 1: Long Papers)",
    month = jul,
    year = "2026",
    address = "San Diego, California, United States",
    publisher = "Association for Computational Linguistics",
    url = "https://aclanthology.org/2026.acl-long.965/",
    doi = "10.18653/v1/2026.acl-long.965",
    pages = "21056--21073",
    ISBN = "979-8-89176-390-6"
}

@inproceedings{
li2025active,
title={Active Evaluation Acquisition for Efficient {LLM} Benchmarking},
author={Yang Li and Jie Ma and Miguel Ballesteros and Yassine Benajiba and Graham Horwood},
booktitle={Forty-second International Conference on Machine Learning},
year={2025},
url={https://openreview.net/forum?id=EHqQaBYYlE}
}

@misc{olmo20252olmo2furious,
      title={2 OLMo 2 Furious}, 
      author={Team OLMo and Pete Walsh and Luca Soldaini and Dirk Groeneveld and Kyle Lo and Shane Arora and Akshita Bhagia and Yuling Gu and Shengyi Huang and Matt Jordan and Nathan Lambert and Dustin Schwenk and Oyvind Tafjord and Taira Anderson and David Atkinson and Faeze Brahman and Christopher Clark and Pradeep Dasigi and Nouha Dziri and Allyson Ettinger and Michal Guerquin and David Heineman and Hamish Ivison and Pang Wei Koh and Jiacheng Liu and Saumya Malik and William Merrill and Lester James V. Miranda and Jacob Morrison and Tyler Murray and Crystal Nam and Jake Poznanski and Valentina Pyatkin and Aman Rangapur and Michael Schmitz and Sam Skjonsberg and David Wadden and Christopher Wilhelm and Michael Wilson and Luke Zettlemoyer and Ali Farhadi and Noah A. Smith and Hannaneh Hajishirzi},
      year={2025},
      eprint={2501.00656},
      archivePrefix={arXiv},
      primaryClass={cs.CL},
      url={https://arxiv.org/abs/2501.00656}, 
}

\clearpage

\appendix

\section{Hyper-parameter Details}
\label{App:hyper}
We set the regularization parameter of Sinkhorn-Knopp algorithm to 0.06, as in \citet{imfeld2024transformer}. We set the global batch size and sequence length to 64 and 2048. For CPT, we use a maximum learning rate of 1e-4 for 1.72B \textit{expanded} models and 5e-5 for 11.5B \textit{expanded} models. For SFT, the maximum learning rate is set to 1e-5 and 5e-6, respectively.

\section{Implementation Details}
\label{App:imp}
We employ Flash-Attention 2 \citep{dao2024flashattention} and  mixed-precision \texttt{bf16} for accelerated training.  We use Language Model Evaluation Harness \citep{eval-harness} for evaluation. 11.5B $expanded$ models are trained on four NVIDIA GH200 (96GB) GPUs while 1.72B $expanded$ models are trained on a single NVIDIA A100 (80GB). We create all $expanded$ models  using AMD EPYC 7413 CPU and a single NVIDIA A100 (80GB).

\section{Performance across checkpoints}
\label{App:performance across checkpoints}

\begin{figure}[!ht] 
\begin{center} 
\includegraphics[width=0.9\linewidth]{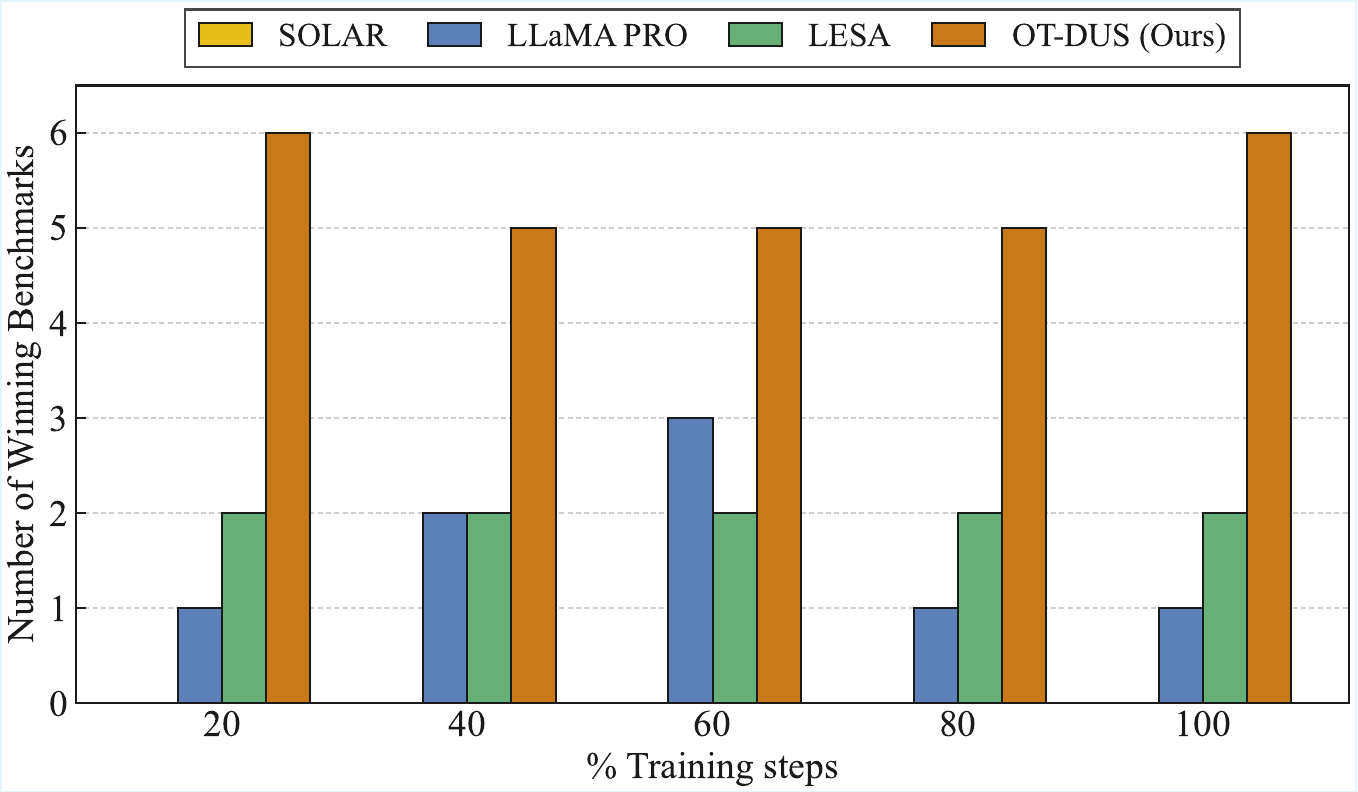} 
\end{center} 
\caption{Number of benchmarks that achieve top performance during the training process of 11.5B \textit{expanded} models. Sums may exceed 8 due to ties.} \label{fig:checkpoint} \end{figure}

\section{Scaling Stability}
\label{App:Up-scale stability}
\begin{table}[!h] 
\centering 
\footnotesize
\setlength{\tabcolsep}{0.35mm} 
\begin{tabular}{lccccc} 
\toprule 
\textbf{ Model} & \textbf{Base} & \textbf{SOLAR} & \textbf{LLaMA PRO} & \textbf{LESA} & \textbf{OT-DUS} \\
 
\midrule  
Llama-3.2-1B      & 11.57 & 16.64 & 11.57 & 871.50 & 11.57 \\
Llama-3.1-8B         & 7.33  & 9.01  & 7.33  & 9.35  & 7.33  \\
\midrule
Mistral-24B & 4.43$^*$  & 6.51$^*$  & 4.43  & 5.17$^*$  & 4.43  \\
Qwen-2.5-32B       & 3.78$^*$  & INF$^*$   & 3.78  & 5.67$^*$  & 3.78  \\
Llama-3-70B        & 1.98$^*$  & 4.21$^*$  & 1.98  & 2.62$^*$  & 1.98  \\
\bottomrule 
\end{tabular}  
\caption{PPL after 1.5x layer expansion initialization for different \textit{base} models, along with PPL of base models. * denotes results from \citet{yang-etal-2025-lesa}, Results for LLaMA PRO and OT-DUS in bottom-half are obtained via reasonable extrapolation from top-half.} 
\label{PPL} 
\end{table}

\section{Additional Computation Cost}
\label{App:Additional Computation Cost}

\begin{table}[!h]
\centering
\footnotesize
\setlength{\tabcolsep}{1mm}
\resizebox{0.5\linewidth}{!}{
\begin{tabular}{@{}lc@{}}
\toprule
\textbf{Expanded Model} & \textbf{Creating Time} \\
\midrule
LESA-1.72B      & 00:26:15 \\
OT-DUS-1.72B    & 00:02:34 \\
\midrule
LESA-11.5B      & 04:52:13 \\
OT-DUS-11.5B    & 00:37:16 \\
\bottomrule
\end{tabular}
}
\caption{Instantiation time for LESA and OT-DUS.}
\label{Extra Time}
\end{table}

Both LESA and OT-DUS require additional computation. LESA necessitates extracting latent patterns using Singular Value Decomposition (SVD), while OT-DUS requires solving the OT problem module-wise. Table \ref{Extra Time} presents the additional computation time required by LESA and OT-DUS to create the \textit{expanded} model. We observe that LESA requires more time compared to OT-DUS (00:26:15 vs. 00:02:34). This time scales massively with larger models (04:52:13 vs. 00:37:16). We hypothesize that this increased time for LESA is mainly caused by the extra computation required for SVD when scaling up base models. Combining Table \ref{Training Time} and Table \ref{Extra Time}, our OT-DUS achieves the best time efficiency among the baselines.

\section{Interpolation Strategy}
\label{App:Interpolation Strategy}

\begin{table*}[!t]
\begin{center}
\footnotesize
\setlength{\tabcolsep}{1mm} 
\resizebox{0.9\textwidth}{!}{
\begin{tabular}{ccccccccccc} 
\toprule
& \multicolumn{1}{c}{\textbf{Perplexity\,$\downarrow$}}
& \multicolumn{8}{c}{\textbf{Zero-shot Performance\,$\uparrow$}} \\ 
\cmidrule(l{2pt}r{2pt}){2-2} \cmidrule(l{2pt}r{2pt}){3-10} 
\textbf{Methods} & \textbf{Wiki-PPL} & \textbf{ARC} & \textbf{LogiQA} & \textbf{Wino} & \textbf{CSQA} & \textbf{BoolQ} & \textbf{PIQA} & \textbf{MMLU} & \textbf{Average} \\ 
\midrule

OT-DUS-Btm  
& $7.83_{0.00}$ 
& $81.75_{0.10}$ 
& \underline{$28.21_{0.09}$} 
& $74.59_{0.41}$ 
& $70.10_{0.14}$ 
& $81.75_{0.02}$ 
& $79.99_{0.13}$ 
& $62.33_{0.08}$ 
& $68.39_{0.12}$ \\

OT-DUS-Mid  
& $\mathbf{7.70}_{\mathbf{0.00}}$ 
& $\mathbf{82.07}_{\mathbf{0.00}}$ 
& $27.65_{0.00}$ 
& $74.32_{0.05}$ 
& \underline{$70.25_{0.23}$} 
& $81.00_{0.12}$ 
& \underline{$80.20_{0.09}$} 
& \underline{$62.53_{0.06}$} 
& $68.28_{0.01}$ \\

OT-DUS-Top  
& \underline{$7.73_{0.00}$} 
& \underline{$82.06_{0.06}$} 
& $27.34_{0.00}$ 
& $\mathbf{74.77}_{\mathbf{0.20}}$ 
& $\mathbf{71.96}_{\mathbf{0.33}}$ 
& $\mathbf{82.18}_{\mathbf{0.07}}$ 
& $\mathbf{80.56}_{\mathbf{0.22}}$ 
& $\mathbf{62.99}_{\mathbf{0.02}}$ 
& $\mathbf{68.84}_{\mathbf{0.04}}$ \\

OT-DUS-T\&B 
& $7.87_{0.00}$ 
& $81.46_{0.10}$ 
& $\mathbf{28.57}_{\mathbf{0.00}}$ 
& \underline{$74.61_{0.18}$} 
& $70.05_{0.05}$ 
& \underline{$82.12_{0.02}$} 
& $79.85_{0.03}$ 
& $62.44_{0.03}$ 
& \underline{$68.44_{0.04}$} \\

\bottomrule
\end{tabular} 
}
\end{center}
\caption{Performance of 11.5B OT-DUS trained on 1.5B tokens using different interpolation strategies.}
\label{tab:opt_deus_variants}
\end{table*}

\section{Ablation Results}
\label{App:ablations}

\begin{table*}[!t]
\begin{center}
\footnotesize
\renewcommand{\arraystretch}{1.25}
\setlength{\tabcolsep}{1mm}
\resizebox{0.95\textwidth}{!}{
\begin{tabular}{c|lccccccccc}
\toprule
\multicolumn{2}{c}{} 
& \multicolumn{1}{c}{\textbf{Perplexity\,$\downarrow$}}
& \multicolumn{7}{c}{\textbf{Zero-shot Performance\,$\uparrow$}} \\
\cmidrule(l{2pt}r{2pt}){3-3}
\cmidrule(l{2pt}r{2pt}){4-10}
\multicolumn{1}{c}{} 
& \multicolumn{1}{c}{\textbf{Methods}}
& \textbf{Wiki-PPL} 
& \textbf{ARC} 
& \textbf{LogiQA} 
& \textbf{Wino} 
& \textbf{CSQA} 
& \textbf{BoolQ} 
& \textbf{PIQA} 
& \textbf{MMLU} 
& \textbf{Average} \\
\midrule

\multirow{2}{*}{\rotatebox[origin=c]{90}{\textbf{TMF}}}
& Attention Output $\mathbf{T}_\text{in}=\mathbf{T}_Q$-1.72B
& $\mathbf{12.05}_{\mathbf{0.00}}$ & $67.12_{0.02}$ & $\mathbf{23.30}_{\mathbf{0.09}}$ & \underline{$60.25_{0.09}$} & $42.64_{0.05}$ & $62.89_{0.04}$ & \underline{$75.39_{0.03}$} & $32.13_{0.05}$ & \underline{$51.96_{0.03}$} \\
& MLP Gate and Up $\mathbf{T}_\text{in}=\frac{1}{2}(\mathbf{T}_O + \mathbf{I})$)-1.72B
& \underline{$12.09_{0.00}$} & $66.81_{0.05}$ & $22.79_{0.18}$ & $\mathbf{60.46}_{\mathbf{0.14}}$ & $42.59_{0.00}$ & $62.22_{0.02}$ & $\mathbf{75.48}_{\mathbf{0.03}}$ & $32.20_{0.07}$ & $51.79_{0.02}$ \\

\midrule

\multirow{3}{*}{\rotatebox[origin=c]{90}{\textbf{OT}}}
& Average-1.72B
& $12.31_{0.27}$ & $67.12_{0.53}$ & $22.38_{0.23}$ & $59.75_{0.50}$ & $41.71_{2.16}$ & $62.76_{0.22}$ & $74.67_{0.14}$ & $32.74_{1.75}$ & $51.59_{0.61}$ \\
& Copy-1.72B
& $12.62_{0.00}$ & $\mathbf{68.00}_{\mathbf{0.02}}$ & $22.73_{0.00}$ & $59.75_{0.14}$ & $34.75_{0.10}$ & \underline{$62.96_{0.02}$} & $74.52_{0.16}$ & $28.20_{0.09}$ & $50.13_{0.01}$ \\
& Random-1.72B
& $16.40_{0.00}$ & $66.79_{0.29}$ & $22.27_{0.00}$ & $59.85_{0.32}$ & $40.95_{0.28}$ & $62.69_{0.05}$ & $74.99_{0.03}$ & $32.19_{0.00}$ & $51.39_{0.13}$ \\

\midrule

\multirow{3}{*}{\rotatebox[origin=c]{90}{\textbf{Zero-Init}}}
& Average + Zero-Init-1.72B
& $12.47_{0.27}$ & \underline{$67.50_{0.54}$} & \underline{$22.89_{0.40}$} & $59.38_{0.72}$ & $40.35_{2.12}$ & $62.54_{0.32}$ & $74.72_{0.08}$ & $31.07_{2.30}$ & $51.20_{0.61}$ \\
& Copy + Zero-Init-1.72B
& $12.23_{0.00}$ & $67.09_{0.07}$ & $22.27_{0.27}$ & $59.48_{0.05}$ & $41.01_{0.24}$ & $\mathbf{64.29}_{\mathbf{0.02}}$ & $75.06_{0.06}$ & $32.45_{0.05}$ & $51.67_{0.00}$ \\
& Random + Zero-Init-1.72B
& $12.20_{0.00}$ & $66.92_{0.00}$ & $21.51_{0.00}$ & $59.27_{0.27}$ & $\mathbf{43.05}_{\mathbf{0.24}}$ & $62.18_{0.02}$ & $74.86_{0.09}$ & $\mathbf{33.29}_{\mathbf{0.05}}$ & $51.58_{0.00}$ \\

\midrule

& OT-DUS-1.72B (Ours)
& $12.19_{0.00}$ & $67.10_{0.09}$ & $22.84_{0.23}$ & $60.06_{0.61}$ & \underline{$42.97_{0.37}$} & $62.75_{0.08}$ & $75.17_{0.14}$ & \underline{$32.88_{0.13}$} & $\mathbf{51.97}_{\mathbf{0.07}}$ \\

\midrule
\midrule

\multirow{2}{*}{\rotatebox[origin=c]{90}{\textbf{TMF}}}
& Attention Output $\mathbf{T}_\text{in}=\mathbf{T}_Q$-11.5B
& $\mathbf{7.47}_{\mathbf{0.00}}$ & $81.80_{0.17}$ & $27.19_{0.00}$ & $73.98_{0.52}$ & \underline{$71.99_{0.41}$} & $82.27_{0.05}$ & \underline{$80.65_{0.06}$} & $\mathbf{63.08}_{\mathbf{0.09}}$ & $68.71_{0.02}$ \\
& MLP Gate and Up $\mathbf{T}_\text{in}=\frac{1}{2}(\mathbf{T}_O + \mathbf{I})$)-11.5B
& $\mathbf{7.47}_{\mathbf{0.00}}$ & $81.80_{0.17}$ & $27.19_{0.00}$ & $74.06_{0.46}$ & $\mathbf{72.02}_{\mathbf{0.41}}$ & $82.31_{0.05}$ & $\mathbf{80.69}_{\mathbf{0.11}}$ & \underline{$63.02_{0.08}$} & $68.73_{0.01}$ \\

\midrule

\multirow{3}{*}{\rotatebox[origin=c]{90}{\textbf{OT}}}
& Average-11.5B
& $7.95_{0.00}$ & $82.07_{0.07}$ & \underline{$27.60_{0.09}$} & $73.69_{0.18}$ & $71.20_{0.09}$ & $82.11_{0.05}$ & $80.23_{0.03}$ & $62.16_{0.04}$ & $68.44_{0.04}$ \\
& Copy-11.5B
& $7.79_{0.00}$ & \underline{$82.21_{0.02}$} & $\mathbf{27.65}_{\mathbf{0.00}}$ & $74.24_{0.09}$ & $71.94_{0.05}$ & $81.60_{0.04}$ & $79.74_{0.13}$ & $62.78_{0.07}$ & $68.59_{0.01}$ \\
& Random-11.5B
& $11.69_{0.00}$ & $79.71_{0.00}$ & $27.19_{0.00}$ & $72.51_{0.04}$ & $69.45_{0.29}$ & $80.57_{0.04}$ & $78.58_{0.06}$ & $61.78_{0.02}$ & $67.11_{0.05}$ \\

\midrule

\multirow{3}{*}{\rotatebox[origin=c]{90}{\textbf{Zero-Init}}}
& Average + Zero-Init-11.5B
& \underline{$7.53_{0.00}$} & $82.14_{0.02}$ & $27.09_{0.09}$ & $74.03_{0.14}$ & $71.83_{0.42}$ & $82.07_{0.02}$ & $80.47_{0.00}$ & $62.70_{0.03}$ & $68.62_{0.03}$ \\
& Copy + Zero-Init-11.5B
& $7.67_{0.00}$ & $\mathbf{82.36}_{\mathbf{0.15}}$ & $27.65_{0.00}$ & $\mathbf{74.87}_{\mathbf{0.05}}$ & $71.64_{0.33}$ & $82.06_{0.02}$ & $80.38_{0.16}$ & $62.80_{0.02}$ & \underline{$68.82_{0.06}$} \\
& Random + Zero-Init-11.5B
& $7.72_{0.00}$ & $81.94_{0.15}$ & $26.42_{0.00}$ & $73.72_{0.00}$ & $71.69_{0.19}$ & $\mathbf{82.47}_{\mathbf{0.02}}$ & $80.02_{0.06}$ & $62.77_{0.01}$ & $68.43_{0.06}$ \\

\midrule

& OT-DUS-11.5B (Ours)
& $7.73_{0.00}$ & $82.06_{0.06}$ & $27.34_{0.00}$ & \underline{$74.77_{0.20}$} & $71.96_{0.33}$ & $82.18_{0.07}$ & $80.56_{0.22}$ & $62.99_{0.02}$ & $\mathbf{68.84}_{\mathbf{0.04}}$ \\

\bottomrule
\end{tabular}
}
\end{center}
\caption{Ablation study on 1.72B and 11.5B \textit{expanded} models.}
\label{tab:ablation_full}
\end{table*}

\begin{table*}[!t]
\begin{center}
\footnotesize
\setlength{\tabcolsep}{1mm} 
\resizebox{0.9\textwidth}{!}{
\begin{tabular}{lcccccccccc} 
\toprule
& \multicolumn{1}{c}{\textbf{Perplexity\,$\downarrow$}}
& \multicolumn{8}{c}{\textbf{Zero-shot Performance\,$\uparrow$}} \\ 
\cmidrule(l{2pt}r{2pt}){2-2} \cmidrule(l{2pt}r{2pt}){3-10} 
\textbf{Methods} & \textbf{Wiki-PPL} & \textbf{ARC} & \textbf{LogiQA} & \textbf{Wino} & \textbf{CSQA} & \textbf{BoolQ} & \textbf{PIQA} & \textbf{MMLU} & \textbf{Average} \\ 
\midrule

Base-8B
& $\mathbf{9.20}_{\mathbf{0.00}}$ 
& \underline{$84.72_{0.15}$} 
& $39.32_{0.00}$ 
& \underline{$73.61_{0.32}$} 
& \underline{$82.25_{0.19}$} 
& $82.27_{0.02}$ 
& $79.47_{0.13}$ 
& $73.47_{0.00}$ 
& $73.59_{0.01}$ \\

SOLAR-11.5B
& $10.47_{0.00}$ 
& $84.25_{0.12}$ 
& $39.32_{0.00}$ 
& $72.11_{0.18}$ 
& $77.01_{0.05}$ 
& $84.56_{0.05}$ 
& $78.44_{0.06}$ 
& $70.84_{0.03}$ 
& $72.36_{0.00}$ \\

LLaMA PRO-11.5B
& \underline{$9.24_{0.00}$} 
& $\mathbf{85.49}_{\mathbf{0.05}}$ 
& $37.94_{0.00}$ 
& $\mathbf{74.11}_{\mathbf{0.00}}$ 
& $81.71_{0.19}$ 
& $84.25_{0.00}$ 
& $79.49_{0.09}$ 
& $73.60_{0.03}$ 
& \underline{$73.80_{0.04}$} \\

LESA-11.5B
& $10.75_{0.00}$ 
& $82.70_{0.07}$ 
& \underline{$40.30_{0.09}$} 
& $73.48_{1.09}$ 
& $81.63_{0.05}$ 
& \underline{$84.80_{0.05}$} 
& \underline{$79.52_{0.03}$} 
& $\mathbf{73.88}_{\mathbf{0.05}}$ 
& $73.76_{0.17}$ \\

OT-DUS-11.5B (Ours)
& $10.58_{0.00}$ 
& $83.38_{0.00}$ 
& $\mathbf{41.83}_{\mathbf{0.09}}$ 
& $73.48_{0.41}$ 
& $\mathbf{84.08}_{\mathbf{0.09}}$ 
& $\mathbf{85.15}_{\mathbf{0.02}}$ 
& $\mathbf{79.54}_{\mathbf{0.19}}$ 
& \underline{$73.84_{0.02}$} 
& $\mathbf{74.47}_{\mathbf{0.06}}$ \\

\bottomrule
\end{tabular} 
}
\end{center}
\caption{CPT Performance of \textit{Expanded} models using Qwen3-8B as \textit{base} model.}
\label{8B_Qwen_Full}
\end{table*}

\end{document}